\pdfoutput=1
\documentclass{article}

\PassOptionsToPackage{numbers}{natbib}
\usepackage[preprint]{neurips_2026}

\usepackage[utf8]{inputenc}
\usepackage[T1]{fontenc}
\usepackage{microtype}
\usepackage{graphicx}
\usepackage{subcaption}
\usepackage{booktabs}
\usepackage{url}
\usepackage{amsmath}
\usepackage{amssymb}
\usepackage{amsfonts}
\usepackage{bm}
\usepackage{mathtools}
\usepackage{amsthm}
\usepackage[table]{xcolor}
\usepackage{wrapfig}
\usepackage{pifont}
\usepackage{nicefrac}

\definecolor{KleinRed}{RGB}{196, 30, 58}
\newcommand{\kred}[1]{\textcolor{KleinRed}{#1}}
\newcommand{\brow}[1]{\rowcolor[HTML]{CFEFFF}#1}

\definecolor{citecolor}{HTML}{0071BC}
\definecolor{linkcolor}{HTML}{ED1C24}

\usepackage[pagebackref=true,breaklinks=true,colorlinks,bookmarks=false,citecolor=citecolor,linkcolor=linkcolor,urlcolor=gray]{hyperref}

\usepackage[capitalize,noabbrev]{cleveref}

\theoremstyle{plain}

\theoremstyle{definition}

\theoremstyle{remark}

\usepackage[textsize=tiny]{todonotes}

\title{MergeTok: Unified Continuous and Discrete Visual Tokenization via Token Merging}

\author{
\begin{tabular}{c}
\textbf{Luyuan Zhang\textsuperscript{$\star$}$^{1}$ \quad Siyuan Li\textsuperscript{$\star$}$^{2,3}$ \quad Zedong Wang$^{4}$ \quad Qingsong Xie \textsuperscript{$\dagger$}$^{5}$ \quad Cheng Tan$^{6}$} \\[0.3em]
\textbf{Anna Wang$^{2}$ \quad Yanhao Zhang$^{5}$ \quad Chen Chen$^{5}$ \quad Haonan Lu$^{5}$ \quad Haoqian Wang\textsuperscript{$\dagger$}$^{1}$} \\[0.5em]
\normalfont
$^1$Tsinghua University \quad
$^2$Westlake University \quad
$^3$Zhejiang University \\[0.1em]
\normalfont
$^4$Hong Kong University of Science and Technology \quad
$^5$OPPO \\[0.1em]
\normalfont
$^6$Shanghai AI Lab \\[0.3em]
\normalfont \textsuperscript{$\star$}Equal contribution. \quad \textsuperscript{$\dagger$}Corresponding author.
\end{tabular}
}

\begin{document}

\maketitle

\begin{abstract}
Most visual tokenizers for image generation are bifurcated into two families with complementary limitations: continuous VAEs offer high-fidelity reconstruction but suffer from dense, entangled latents that are poorly suited for semantic control, whereas discrete VQ-based models enable autoregressive generation yet struggle with gradient sparsity, unstable training, and codebook collapse. In this work, we introduce MergeTok, a unified tokenizer that jointly optimizes continuous (VAE) and discrete (VQ) tokenizers within a encoder-decoder architecture, leveraging token merging techniques as a semantic bridge. By clustering similar tokens during encoding, MergeTok establishes a structural prior that provides dual supervision signals: (i) it imposes merged-token semantic alignment in the VAE branch, regularizing its latent space toward disentangled, semantic-aware representations; (ii) it derives group-wise constraints, promoting intra-group diversity and inter-group exclusivity that stabilize VQ training. MergeTok shows competitive reconstruction and generation performance on ImageNet-256, with substantially lower rFID than strong VAE and VQ models under matched token budgets, while producing semantically-organized token representations compatible with both autoregressive and diffusion generators. This shows that a single architecture can endow visual tokenizers with robust semantic organization and generator-friendly discreteness.
\end{abstract}

\section{Introduction}

Image generation has achieved remarkable success with both auto-regressive~\cite{cvpr2021vqgan, Zhang2025V2FlowUV} and diffusion-based paradigms~\cite{iccv2023DiT}. These advances rely heavily on the underlying visual tokenizer, which is typically implemented as an auto-encoder that maps an image to a sequence of latent features and imposes a \textit{latent-space constraint} to encode them either as continuous variables or discrete codes. Existing tokenizers are largely divided by the form of this latent constraint. \textbf{VQ-based} tokenizers such as VQ-VAE~\cite{2017VQ-VAE} quantize encoder outputs into discrete codebook indices, which naturally support sequence modeling with categorical distributions. \textbf{VAE-based} tokenizers, in contrast, map features to a continuous probabilistic latent space and favor reconstruction fidelity together with stable gradient propagation~\citep{iclr2013VAE, cvpr2025VAVAE}. It is widely believed that advances in the encoder and latent design of tokenizers are central to downstream generation, since these components largely determine token efficiency and semantic controllability~\citep{Kim2025TATiTok, cvpr2025MergeVQ}.

\begin{figure}[t!]
    \centering
    \includegraphics[width=1.0\linewidth]{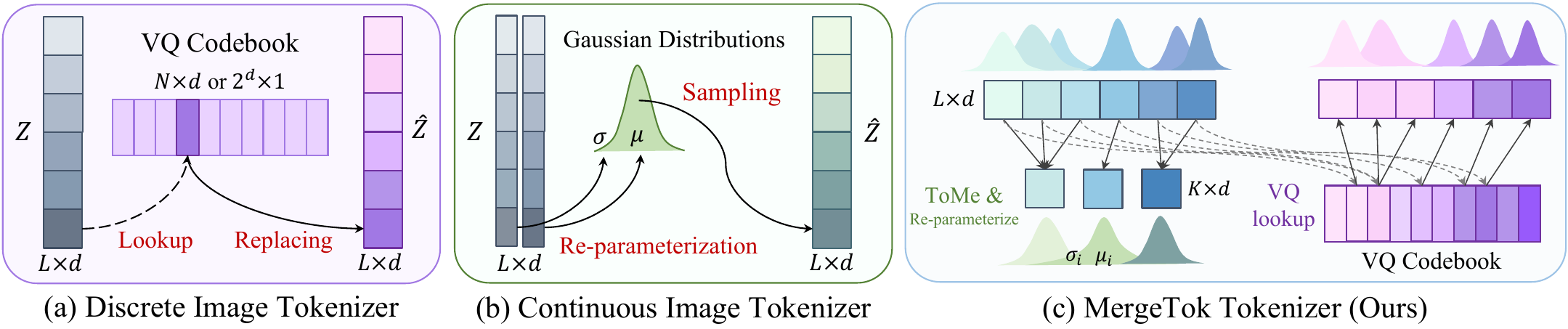}
    \caption{
    \textbf{(a) Discrete VQ.} Features are quantized by nearest-neighbor codebook lookup, but codebook updates can be sparse.
    \textbf{(b) Continuous VAE.} Features are mapped to continuous Gaussian latents through reparameterization for stable reconstruction.
    \textbf{(c) MergeTok.} MergeTok adopts a dual-branch design to jointly optimize VAE and VQ tokenization. The VAE branch introduces online token merging to inject semantic structure into continuous latents, while the resulting token-similarity information is used to guide group-aware VQ quantization, improving discrete token learning.
}
    \label{fig:mergetok_intro}
\end{figure}
Despite recent progress, both paradigms remain limited by structural weaknesses. \textbf{(a)} \textit{Continuous VAE-based tokenizers} rely on a latent space that lacks intrinsic organization. Under strong compression, the encoder is driven to preserve fine-grained visual information in dense latent representations, where multiple semantic factors become entangled across latent components. As a result, the learned codes are often weakly disentangled and poorly factorized, which restricts controllable generation and makes it difficult to obtain compact representations that are both semantically structured and reconstruction-friendly. \textbf{(b)} \textit{discrete VQ tokenizers} face a different bottleneck rooted in optimization. Since quantization is non-differentiable, training depends on approximate gradient propagation, which produces sparse and uneven updates to the codebook. Many codebook entries therefore receive little supervision, increasing the risk of collapse and reducing representation efficiency. These issues create a persistent tension among reconstruction fidelity, semantic structure, efficiency, and training stability.

To overcome these limitations, we propose MergeTok, a unified tokenizer that jointly optimizes a VAE branch and a VQ branch, as illustrated in Fig.~\ref{fig:mergetok_intro} (c). Our key observation is that token merging provides a natural semantic interface between continuous and discrete tokenization: when content-similar tokens are merged into groups, the resulting source map $S$ can be reused as a structural prior for both branches. We exploit this in two complementary ways. First, we impose merged-token alignment to regularize the VAE latent space toward semantic-aware representations. Second, we use the induced clustering to guide group-aware VQ quantization, promoting within-group code diversity and cross-group exclusivity that stabilize codebook learning. Both branches share the encoder and decoder, are trained jointly under a unified objective, and output 256 tokens at the tokenizer interface. ToMe operates entirely inside the tokenizer's training loop and is invisible to downstream generators, as detailed in Sec.~\ref{sec3.2_framework}. Experiments on ImageNet-256 demonstrate strong rFID and perceptual quality, consistently outperforming strong VAE-Only and VQ-Only baselines.

Our contributions are summarized as follows:
\begin{itemize}
    \item \textbf{A unified dual-branch tokenizer.} We propose \emph{MergeTok}, a tokenizer that jointly optimizes \emph{continuous} and \emph{discrete} latents under a single objective. By using token merging as the semantic connection between the two branches, the stability of VAE training and the structural advantages of VQ can reinforce each other in one encoder-decoder framework.
    \item \textbf{Merge-aware training constraints.} We introduce two simple yet effective objectives derived from the merging process: \emph{merged-token alignment}, which improves semantic structure in the continuous latent space, and \emph{group-aware quantization}, which stabilizes VQ training and improves codebook utilization.
    \item \textbf{Granularity-aware merge ratio sampling.} We employ discrete merge-ratio sampling during training so that the model is exposed to multiple token granularities. This enables unified tokenization with improved fidelity and efficiency across reconstruction and generation.
\end{itemize}

\section{Related Work}
\label{sec:related_work}

\paragraph{Visual Tokenizers for Image Generation.}
Modern visual tokenizers convert images into discrete or continuous sequences to enable transformer-based generation. On the discrete side, VQ-based methods address gradient sparsity and codebook collapse through improved differentiability and scaling. IBQ~\cite{Shi2024IBQ} propagates gradients over full code distributions to maintain high code utilization, while LFQ~\cite{CVPR2023MAGVIT} replaces vector lookups with binary indices to support very large vocabularies. SoftVQ-VAE~\cite{cvpr2025SoftVQVAE} adopts soft categorical posteriors for differentiable training and strong compression. On the continuous side, DC-AE~\cite{iclr2025DCAE} combines residualized latents with staged training to sustain quality at extreme spatial downsampling, and REPA~\cite{Yu2024REPA} aligns latent features to visual encoders to stabilize diffusion optimization and improve generative quality. Collectively, these advances expand vocabulary capacity, restore gradient flow, and impose semantic structure on visual representations.

\paragraph{Unified Discrete and Continuous Visual Tokenizers.}
A growing line of work produces both discrete and continuous representations within a single tokenizer to leverage their complementary strengths. Wave-Particle~\cite{Chen2025DCC-VT} builds two parallel branches, a continuous VAE branch for reconstruction and a discrete VQ branch for generation, with partial parameter sharing. VAEVQ~\cite{Yang2025VAEVQED} introduces a variational prior into codebook learning so that discrete tokens inherit the regularity of continuous latent structure. TokenBridge~\cite{iccv2025TokenBridge} approaches unification from the generation side, encoding images into continuous tokens and subsequently discretizing them for autoregressive synthesis. MergeTok shares this goal but uses the token-merging source map as a structural interface coupling both branches through a shared encoder and decoder, serving as an alignment signal for the continuous branch and a grouping prior for the discrete branch. A detailed design comparison is provided in Appendix~\ref{app:related}.

\paragraph{Unified Visual Tokenizers for Multimodal and Multi-Task Systems.}
Beyond generation, recent work designs tokenizers that serve both recognition and synthesis across multiple modalities. UniTok~\cite{Ma2025UniTok} shows that multi-codebook quantization and wider embeddings mitigate the reconstruction-semantics conflict, yielding low rFID and strong zero-shot accuracy while enabling native generation in MLLMs. AToken~\cite{Lu2025ATokenAU} unifies images, video, and 3D in a single token space via a transformer with 4D positional encoding, achieving high-fidelity reconstruction and competitive downstream performance. SPAE~\cite{nips2023SPAE} aligns vision with language by translating images into multi-scale lexical tokens consumable by frozen LLMs, enabling both understanding and token-level image synthesis. Similarly, MAGVIT-v2~\cite{CVPR2023MAGVIT} shows that with better visual tokens, GPT-style generators can rival or surpass diffusion models on standard benchmarks. These works illustrate that capacity scaling, multi-scale design, and language alignment enable a single tokenizer to support diverse tasks and modalities.

\paragraph{Token Compression Techniques and Their Impact on Generative Models.}
A complementary line of work reduces sequence length to improve efficiency and robustness, particularly for autoregressive decoders where errors accumulate with depth. Adaptive-length tokenization~\cite{nips2024titok} distills 2D patches into compact 1D sequences by allocating tokens according to content entropy, reducing unnecessary decoding steps. Fixed aggressive compression via tokenizers such as SoftVQ-VAE~\cite{cvpr2025SoftVQVAE} and DC-AE~\cite{iclr2025DCAE} achieves tens of tokens per image with competitive fidelity. At inference time, ToMe~\cite{iclr2022ToMe} dynamically fuses redundant tokens to reduce compute and memory with negligible perceptual loss. Training-time designs such as MergeVQ~\cite{cvpr2025MergeVQ} integrate merging into the encoder to form short, quantized sequences while retaining mechanisms to recover fine detail at decode time. Together, adaptive token counts and principled merging reduce step count and memory pressure while limiting error propagation. In contrast, MergeTok uses merging as a structural mechanism, with the source map serving as an interface between the continuous and discrete branches rather than as a compression strategy.

\section{Methods}
\subsection{MergeTok Framework}
\label{sec3.2_framework}

\begin{figure}[t!]
    \centering
    \includegraphics[width=1.0\linewidth]{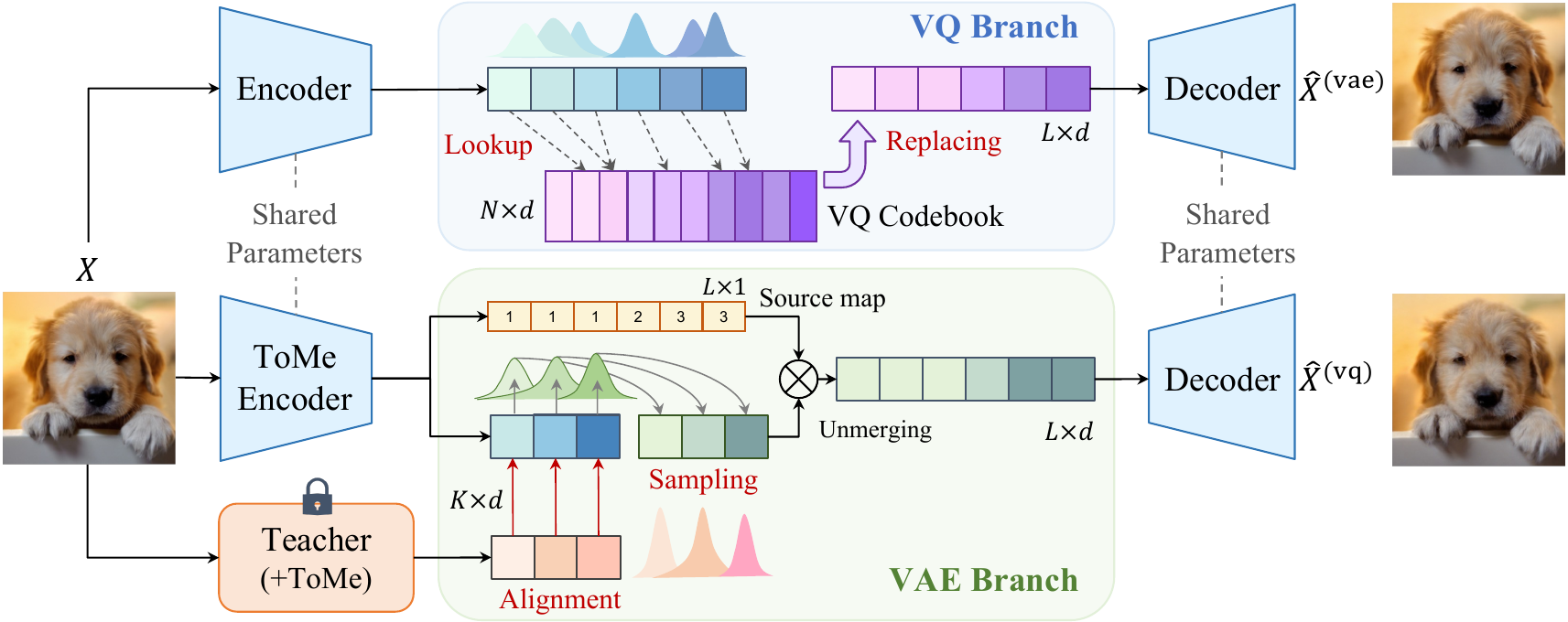}
    \caption{\textbf{Overall Framework of MergeTok.} We propose a dual-branch architecture that jointly optimizes continuous and discrete representations with shared encoder and decoder. \textbf{(i) VAE Branch (Bottom)} applies ToMe~\citep{iclr2022ToMe} to extract dense semantic tokens, which are aligned with a teacher model (also equipped with ToMe). The resulting source map is then employed to unmerge the groups back to the full lattice for reconstruction. \textbf{(ii) VQ Branch (Top)} inherits this source map to induce group-aware clustering, enforcing intra-group diversity and inter-group exclusivity constraints that stabilize the training of the discrete codebook.
    }
    \label{fig:framework}
\end{figure}

The continuous (VAE) and discrete (VQ) paradigms of visual tokenization are usually treated as alternatives, each with its own well-known weakness --- entangled latents on the continuous side, and gradient sparsity with codebook collapse on the discrete side. Our key observation is that these two weaknesses are addressable by the \emph{same} structural signal: a soft clustering of visually similar tokens. If such a clustering is available, it can regularize the continuous latent toward semantic groups while simultaneously partitioning the discrete codes for group-aware quantization. MergeTok turns this observation into an architecture: a dual-branch tokenizer in which token merging (ToMe) supplies the shared structural signal, as illustrated in Fig.~\ref{fig:framework}.

Given an input image \( X \in \mathbb{R}^{H \times W \times 3} \), we encode it through a shared CNN encoder \( \mathcal{E}_c(\cdot) \), producing a feature map \( Z \in \mathbb{R}^{\frac{H}{f} \times \frac{W}{f} \times D} \), where \( f \) is the downsampling factor and \( D \) the channel dimension. The map is reshaped into a token sequence \( Z_L = \mathcal{E}_c(X) \in \mathbb{R}^{L \times D} \) with \( L = \frac{H}{f} \cdot \frac{W}{f} \), which serves as input to both the VAE and VQ branches.

\paragraph{VAE Branch.} The token sequence \(Z_L\) is fed into an attention-based encoder \( \mathcal{E}_a(\cdot; r) \) equipped with token-merging modules for further feature extraction, where \(r\in(0,1]\) controls the token keep ratio over \(N\) layers. This produces a condensed representation \( Z_K^{(\mathrm{vae})} \in \mathbb{R}^{K \times D} \) with \(K=\lfloor \kappa L \rfloor\) (e.g., \(\kappa=r^N\)), together with a binary source map \(S\) that records the assignment from the \(L\) pre-merge tokens to the \(K\) merged ones:
\begin{equation}
\label{equ:vae_forward}
    Z_K^{(\mathrm{vae})},\; S \;=\; \mathcal{E}_a\!\left(Z_L;\, r, N\right).
\end{equation}
For reconstruction, we employ a hybrid VAE decoder \( \mathcal{D}^{(\mathrm{vae})}(\cdot) \) that jointly takes the merged tokens and the source map to recover pixel-space details, yielding \(\hat{X}^{(\mathrm{vae})} = \mathcal{D}^{(\mathrm{vae})}\!\left(Z_K^{(\mathrm{vae})},\, S\right)\).
We defer the implementation details to Sec.~\ref{sec3.3_vae}.

\paragraph{VQ Branch.} The VQ branch uses the same attention encoder \( \mathcal{E}_a(\cdot) \) with the ToMe operation bypassed (equivalent to keeping all $L$ tokens), preserving the full-length sequence:
\begin{equation}
\label{eq:vq_forward}
    Z_L^{(\mathrm{vq})} \;=\; \mathcal{E}_a\!\left(Z_L;\, r{=}0 \right)\in\mathbb{R}^{L\times D}.
\end{equation}
We quantize each token \(z_t\in Z_L^{(\mathrm{vq})}\) with a codebook \(\mathcal{C}=\{c_j\}_{j=1}^{n}\) via:
\begin{equation}
\label{equ:quantize}
    \tilde{z}_t \;=\; \mathcal{Q}(z_t)\;=\; c_{i^\ast},\quad
    i^\ast \;=\; \arg\min_{j\in\{1,\dots,n\}} \|z_t-c_j\|_2,
\end{equation}
yielding \(\tilde{Z}_L^{(\mathrm{vq})}=\{\tilde{z}_t\}_{t=1}^{L}\). During quantization, we further leverage the source map \(S\) produced by the VAE branch to impose group-aware guidance on code assignments and improve codebook learning. Finally, a hybrid VQ decoder reconstructs the image as \(\hat{X}^{(\mathrm{vq})} = \mathcal{D}^{(\mathrm{vq})}\!\left(\tilde{Z}_L^{(\mathrm{vq})}\right)\).
We defer the computation details to Sec.~\ref{sec3.4_vq}.

\paragraph{The Role of Token Merging in MergeTok.}
Concretely, the shared structural signal takes the form of a source map $S$ produced by ToMe inside the encoder. Unlike prior work that uses ToMe for inference-time acceleration or output-sequence compression~\cite{iclr2022ToMe}, MergeTok integrates ToMe within the tokenizer's \emph{training loop}: in the VAE branch, $S$ unmerges the $K$ condensed tokens back to the full $L$-token lattice for reconstruction, while $Z_K^{(\text{vae})}$ is used for teacher alignment (Eq.~\ref{eq:loss_align}); in the VQ branch, ToMe is bypassed and $S$ enters only as a regularization signal on codebook usage statistics (Eqs.~\ref{eq:loss_div}--\ref{eq:loss_cons}). Consequently, the tokenizer outputs 256 tokens, and downstream generators such as LlamaGen, DiT, and SiT are agnostic to the internal merging structure.

\subsection{Semantic Enhancement in VAE Branch}
\label{sec3.3_vae}
\begin{figure}[t!]
    \centering
    \includegraphics[width=1.0\linewidth]{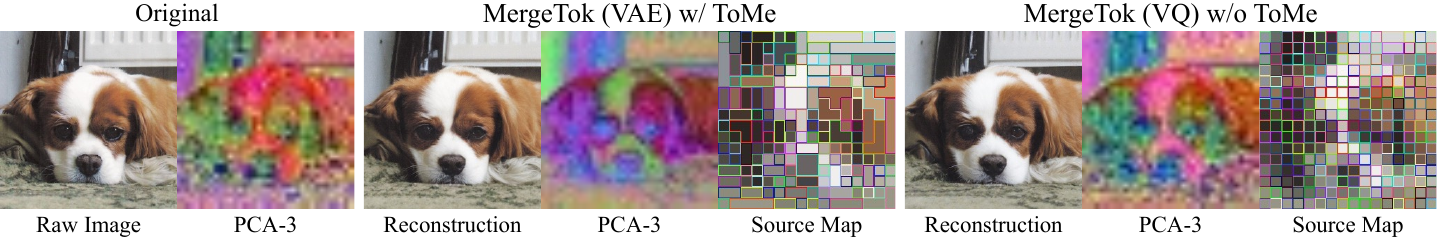}
    \caption{\textbf{Semantic Condensing Effects.}
We visualize PCA-3 components of raw/reconstructed images and the corresponding ToMe source maps to show how MergeTok organizes visual information.
The VAE branch is constrained by token-wise aggregation, yielding semantic separability comparable to discrete models.
The VQ branch without ToMe shows inherent clustering due to quantization.
}
    \label{fig:semantic}
\end{figure}

\paragraph{Token Merging for Semantic Abstraction.}
We apply the token merging algorithm in ToMe as the fusion module in our attention encoder \(\mathcal{E}_a(\cdot)\).
Concretely, \(\mathcal{E}_a(\cdot)\) compresses the input sequence by merging the most similar tokens at each layer, controlled by a prescribed per-layer keep ratio \(r\in(0,1]\) over \(N\) merging layers. Denoting the effective keep ratio by \(\kappa=r^N\) (or more generally \(\kappa=\prod_{\ell=1}^{N} r_\ell\) for layer-wise ratios \(\{r_\ell\}\)), the encoder outputs a condensed \(K\)-token representation \(Z_K^{(\mathrm{vae})}\in\mathbb{R}^{K\times D}\) with \(K=\lfloor \kappa L\rfloor\), together with a source map \(S \in \{1,\dots,K\}^{1\times L}\) that records the assignment from original to merged tokens: the \(i\)-th entry \(S[i]\) specifies the merged index, \(S[i]=k\) means the original token \(i\) is merged into the \(k\)-th token in \(Z_K^{(\mathrm{vae})}\).
For reconstruction, we convert the source map \(S\) into a one-hot assignment matrix \(A\in\{0,1\}^{L\times K}\) as:
\begin{equation}
\label{equ:source_matrix}
A_{i,k} \;=\; \mathbf{1}[S[i]=k],
\qquad A\in\{0,1\}^{L\times K}.
\end{equation}
We then restore the original token layout via:
\begin{equation}
\tilde{Z}_L \;=\; A\, Z_K^{(\mathrm{vae})} \in \mathbb{R}^{L\times D}.
\end{equation}
The recovered sequence \(\tilde{Z}_L\) is decoded into the pixel-space \(\hat{X}^{(\mathrm{vae})} = \mathcal{D}^{(\mathrm{vae})}\!\left(Z_K^{(\mathrm{vae})},\, S\right)\).
A more detailed computational example is provided in Appendix~\ref{app:tome}.

As such, the VAE branch discovers sample-wise semantic clusters at encoding time, decoupling high-frequency details and thereby preventing an overly dense latent space.

\paragraph{Semantic Alignment at Matched Granularity.}
To regularize the VAE latent space, we align merged tokens with a frozen DINO-style teacher \( \mathcal{T} \) configured with the same merging schedule \((r,N)\). The teacher produces \(K\) semantic features \(F_K^{(\mathrm{tea})} \in \mathbb{R}^{K \times D_t}\), and we project the merged latents \(Z_K^{(\mathrm{vae})}\in \mathbb{R}^{K \times D_t}\) via an alignment head \(\mathcal{H}_{\mathrm{ali}}\):
\begin{equation}
F_K^{(\mathrm{tea})}=\mathcal{T}(X;\,r,N),
\qquad
\bar{Z}_K=\mathcal{H}_{\mathrm{ali}}\!\left(Z_K^{(\mathrm{vae})}\right).
\end{equation}
We then compute the alignment loss \( \mathcal{L}_{\mathrm{align}} \) between the projected student features \( \bar{Z}_K \) and the teacher features \( F_K^{(\mathrm{tea})} \) using a similarity metric as cosine distance:
\begin{equation}
\label{eq:loss_align}
    \mathcal{L}_{\mathrm{align}} = \frac{1}{K} \sum_{k=1}^{K} \left\| \bar{Z}_K[k] - F_K^{(\mathrm{tea})}[k] \right\|_2^2,
\end{equation}
where \( \bar{Z}_K[k] \) and \( F_K^{(\mathrm{tea})}[k] \) are the \( k \)-th merged token from the student and teacher, respectively.
\begin{figure}[t!]
    \centering
    \includegraphics[width=0.99\linewidth]{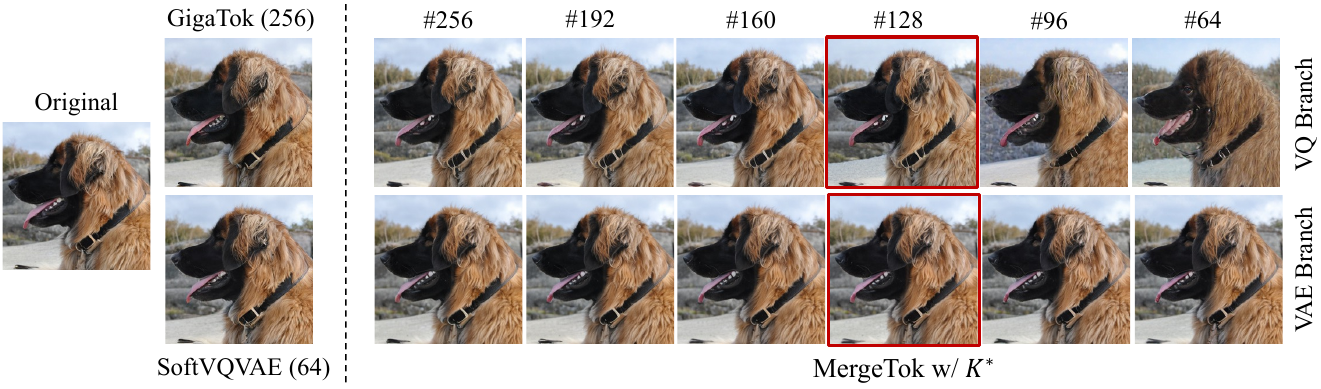}
    \caption{\textbf{Reconstruction in both VQ and VAE branches across Token Granularities.} We visualize reconstructions from both branches while sweeping the target sampling center $K^*$ controlled by merge ratio $r$ from \#256 (left) to \#64 (right). It shows MergeTok's robustness to varying compression rates. The \kred{\textbf{red marker}} indicates the optimal kept-token count ($K^*=128$) found during training.
}
    \label{fig:rec_vis}
\end{figure}

This alignment loss encourages VAE latent tokens to capture  meaningful structure aligned with a teacher model. Rather than aligning only the [CLS] token, which overlook fine-grained semantics, or all patch tokens, which can be overly rigid, we align at the semantic level using merged tokens. This facilitates focus on global semantics while reducing sensitivity to irrelevant details, leading to more coherent latent representations for generation as shown in Fig.~\ref{fig:semantic}.

\subsection{Improving VQ with VAE-Derived Group Priors}
\label{sec3.4_vq}

The VAE branch improves VQ along two axes: (1) joint training routes continuous gradients through the shared encoder, alleviating the biased straight-through estimates of quantization (Eq.~\ref{equ:quantize}) and yielding more stable optimization (Appendix~\ref{app:ablation_component}--\ref{app:codebook_health}); (2) the VAE-derived source map $S$ serves as an online clustering prior that partitions the $L$ VQ tokens into $K$ semantic groups, enabling two group-aware regularizers: $\mathcal{L}_{\mathrm{div}}$ for intra-group diversity and $\mathcal{L}_{\mathrm{cons}}$ for inter-group exclusivity.

As in Eq.~\ref{equ:vae_forward} and Eq.~\ref{equ:source_matrix}, the VAE branch provides a source map \(S\) and its one-hot variant \(A\in\{0,1\}^{L\times K}\), where \(A_{i,g}=1\) indicates that the \(i\)-th original token belongs to group \(g\).
Let \(q_i\in\{1,\dots,n\}\) be the hard code index assigned to the \(i\)-th VQ token (codebook size \(n\)). As such, the group size is $N_g = \sum_{i=1}^{L} A_{i,g}$.
We then summarize code usage \emph{within each group} by a categorical distribution \(p_g\in\mathbb{R}^{n}\), whose \(k\)-th entry counts the fraction of tokens in group \(g\) assigned to code \(k\):
\begin{equation}
p_{g,k}
\;=\;
\frac{1}{N_g}\sum_{i=1}^{L} A_{i,g}\,\mathbf{1}[q_i=k],
\qquad k\in\{1,\dots,n\}.
\label{eq:pgk}
\end{equation}
Since \(\sum_{k=1}^{n}\mathbf{1}[q_i=k]=1\) for each token \(i\), we have \(\sum_{k=1}^{n}p_{g,k}=1\), so \(p_g\) is a valid probability distribution over code indices for group \(g\).

\textit{Intra-group Diversity Loss.}
To prevent a semantic group from collapsing to a single code, we maximize the entropy of \(p_g\). Equivalently, we minimize the negative entropy:
\begin{equation}
\mathcal{L}_{\mathrm{div}}
\;=\;
-\sum_{g=1}^{K} H(p_g)
\;=\;
\sum_{g=1}^{K}\sum_{k=1}^{n} p_{g,k}\log p_{g,k}.
\label{eq:loss_div}
\end{equation}
where \(H(p_g)=-\sum_{k}p_{g,k}\log p_{g,k}\) is maximized when codes are used more evenly within group \(g\), thus encouraging \emph{diverse-within} code assignments and improving codebook utilization.

\textit{Inter-group Consistency Loss.}
While \(\mathcal{L}_{\mathrm{div}}\) encourages diversity within each group, we additionally enforce \emph{separation between groups} by discouraging different groups from using the same codes.
We measure the overlap between groups \(g\) and \(h\) by dot product of code-usage distributions: $\langle p_g, p_h\rangle = \sum_{k=1}^{n} p_{g,k}\,p_{h,k}$.
This quantity is large when both groups place mass on the same set of codes, and small when they use disjoint code subsets.
Aggregating this overlap over all group pairs yields:
\begin{equation}
\mathcal{L}_{\mathrm{cons}}
\;=\;
\sum_{\substack{g,h=1\\ g\neq h}}^{K}\;\sum_{k=1}^{n} p_{g,k}\,p_{h,k}
\;=\;
\sum_{\substack{g,h=1\\ g\neq h}}^{K} \langle p_g, p_h\rangle.
\label{eq:loss_cons}
\end{equation}
Minimizing \(\mathcal{L}_{\mathrm{cons}}\) therefore drives different groups to specialize on different code subsets, yielding \emph{exclusive-between} code usage and clearer inter-group separation.

Together, these regularizers improve codebook utilization and prevent codebook collapse; corresponding statistics are reported in Appendix~\ref{app:codebook_health}.

\subsection{Training Strategies}
\label{Sec3.4}

\paragraph{Total Learning Objective.}
We follow the classical training recipes of VAE and VQ tokenizers in our two branches. The VAE branch uses a standard objective $\mathcal{L}_{\mathrm{VAE}}$ that combines pixel reconstruction $\mathcal{L}_{\mathrm{vae\mbox{-}rec}}$, perceptual similarity $\mathcal{L}_{\mathrm{perc}}$, a KL regularizer $\mathcal{L}_{\mathrm{KL}}$ on $q_\varphi(Z_K^{(\mathrm{vae})}\!\mid X)$, and an adversarial term $\mathcal{L}_{\mathrm{gan}}^{G}$. The VQ branch is supervised by $\mathcal{L}_{\mathrm{VQ}}$ that combines reconstruction $\mathcal{L}_{\mathrm{vq\mbox{-}rec}}$, codebook update $\mathcal{L}_{\mathrm{codebook}}$, commitment $\mathcal{L}_{\mathrm{com}}$, and adversarial $\mathcal{L}_{\mathrm{gan}}^G$ losses. The total objective combines the two branches with our three merge-aware regularizers:
\begin{equation}
\label{eq:total_simple}
\begin{aligned}
\mathcal{L}_{\mathrm{total}}
~=~
&\lambda_{\mathrm{vae}}\mathcal{L}_{\mathrm{VAE}}
~+~
\lambda_{\mathrm{vq}}\mathcal{L}_{\mathrm{VQ}}
 ~+~
\lambda_{\mathrm{ali}}\mathcal{L}_{\mathrm{align}}
 ~+~
\lambda_{\mathrm{div}}\mathcal{L}_{\mathrm{div}}
~+~
\lambda_{\mathrm{cons}}\mathcal{L}_{\mathrm{cons}},
\end{aligned}
\end{equation}
where $\mathcal{L}_{\mathrm{align}}$, $\mathcal{L}_{\mathrm{div}}$, $\mathcal{L}_{\mathrm{cons}}$ are defined in Eqs.~(\ref{eq:loss_align})--(\ref{eq:loss_cons}); loss weights are detailed in Sec.~\ref{sec:exp}.

\paragraph{Dynamic Sampling of Merge Ratios.}
To improve the robustness of the VAE decoder to varying token granularities, we perform dynamic sampling to the token merging ratio during training. Specifically, at each training step, we discretely sample the number of retained tokens \(K \in \{k_1,k_2 \dots, k_t\}\) from a truncated Gaussian distribution with mean $\mu$ and standard deviation \(\sigma\), written as:
\begin{equation}
    k \sim \text{Discrete-}\mathcal{N}(\mu = K^*, \sigma^2),\quad \text{s.t. } k \in \{k_1, \dots, k_t\},
\end{equation}

\begin{wrapfigure}[14]{r}{0.46\textwidth}
\centering
\vspace{-1.25em}
\includegraphics[width=1.0\linewidth]{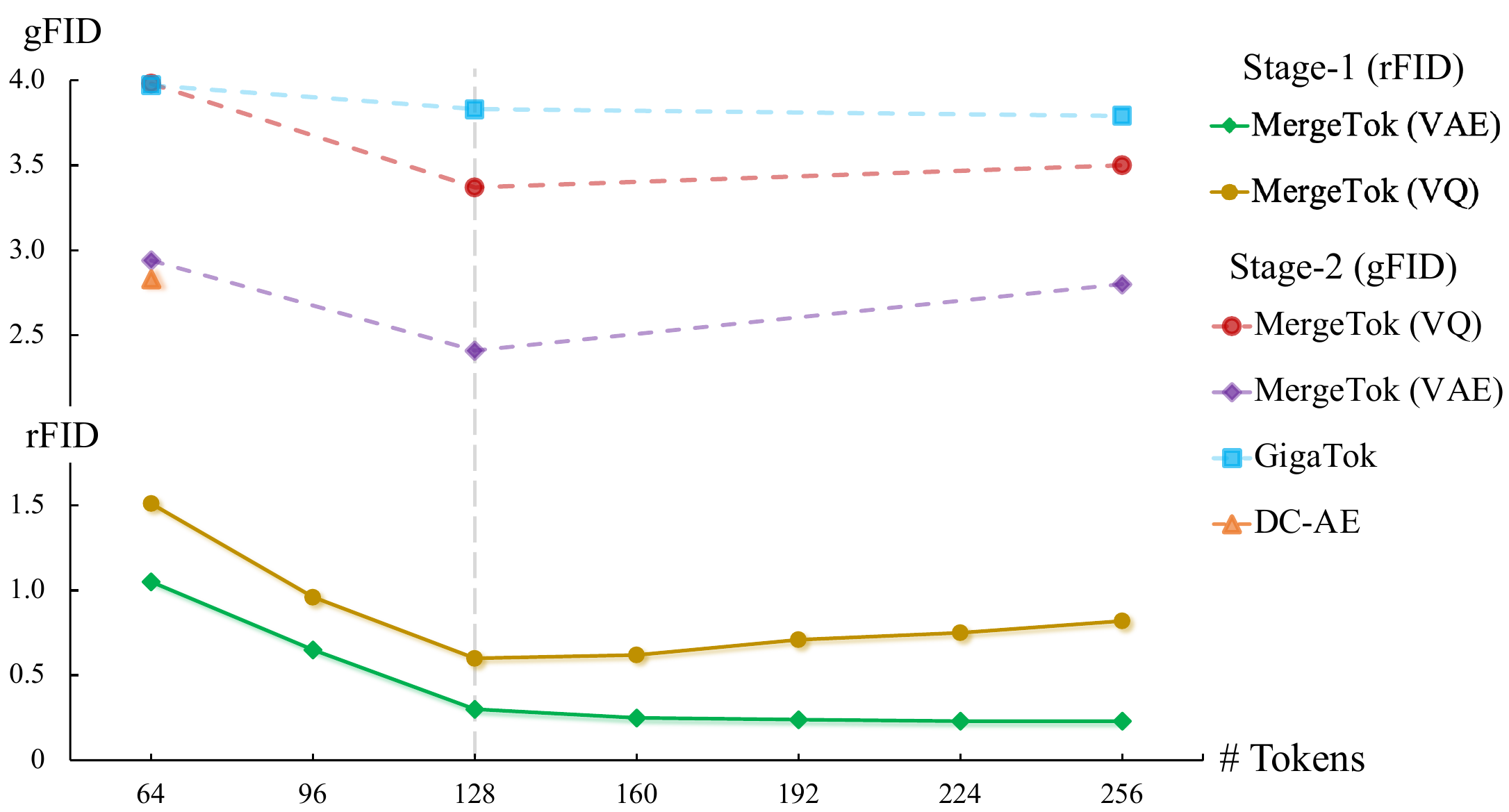}
\vspace{-0.75em}
\caption{\textbf{Kept token number \textit{vs} rFID/gFID.} With $K^* = 128$, MergeTok achieves competitive rFID and gFID.}
\vspace{-4.0em}
\label{fig:token_vs_fid}
\end{wrapfigure}

\noindent where $K^*$ denotes a hyperparameter that approximates the dataset's \emph{information density}, and $\{k_i\}$ enumerates the admissible kept-token counts; $\sigma$ controls the dispersion of the discrete Gaussian, and sampling is clipped to the valid set. The corresponding merge ratio is then computed by
a scheduling function, $r = \mathrm{merge\_ratio\_generator}(K, \beta)$, which determines the merge ratio \(r\) given the retained token count \(K\) and a decay factor \(\beta\). This exposes the encoder to varied internal bottleneck widths during training, encouraging it to organize information that remains recoverable across multiple semantic granularities via the unmerge operation. As discussed in Sec.~\ref{sec3.2_framework}, this dynamic sampling acts purely as a training-time regularizer; the tokenizer always outputs 256 tokens. We show the reconstruction result of the VQ and VAE branch with different sampled merge ratios in Fig.~\ref{fig:rec_vis}, and the rFID and gFID of different methods with different remaining tokens in Fig.~\ref{fig:token_vs_fid}.

\section{Experiments}
\label{sec:exp}

\subsection{Experimental Protocol}
\label{sec:exp_protocol}

\paragraph{Evaluation axes.}
We design our experiments to evaluate MergeTok along three axes: \emph{reconstruction fidelity}, \emph{representation quality}, and \emph{downstream generation}. This protocol directly tests whether a single tokenizer can serve both discrete autoregressive generation and continuous diffusion/flow generation while inducing semantic organization in the latent space, which is the central claim of a unified tokenizer.

\paragraph{Tokenizer setup.}
We instantiate two parameterizations of MergeTok, denoted SB and BL. The SB variant uses a 19M-parameter attention encoder (6 blocks, 8 heads, dim 512) and an 86M-parameter decoder (6 blocks, 12 heads, dim 768). The BL variant scales the encoder to 86M and the decoder to 329M (24 blocks, 16 heads, dim 1024). A frozen DINOv2 ViT-B serves as the alignment teacher, and the VQ branch uses a codebook of size $16{,}384$ with $8$-dimensional code embeddings. During training, the merged-token count is sampled from $K\in\{96, 128, 160, 192, 224, 256\}$ centered at $K^{*}=128$; at evaluation, $K=K^{*}$ for both branches and the tokenizer always emits 256 tokens. We emphasize that ToMe is an internal training-time mechanism of the tokenizer: downstream generators operate directly on the 256-token sequence without any merge-related modification, ensuring a controlled comparison against non-merge tokenizers. All MergeTok models are trained on ImageNet-1K at $256{\times}256$ for 200 epochs with batch size 256 using AdamW (lr $1\text{e-}4$, cosine decay, $\beta_1{=}0.9$, $\beta_2{=}0.95$). The total objective weights the VQ, VAE, alignment, intra-group diversity, and inter-group consistency losses with coefficients $1.0$, $1.0$, $0.5$, $0.05$, $0.05$, respectively.

\begin{table}[t]
    \centering
    \caption{\textbf{System-level comparison of discrete tokenizers on ImageNet 256$\times$256}. We report rFID for reconstruction, linear probing accuracy (Lin.) for representation, and gFID/IS for class-conditional generation. $\star$: trained with frozen DINO discriminator. ``CFG": classifier-free guidance.
    }
    \setlength{\tabcolsep}{1.2pt}
    \resizebox{1.0\linewidth}{!}{
    \footnotesize
    \begin{tabular}{l|cccccc|ccc|cc|cc}
    \toprule
Tokenizer                                            &       & \multicolumn{3}{c}{VQ Codebook}    & Lin.      & rFID      & Generator              & \#Param. & \#Step & \multicolumn{2}{c|}{w/o CFG}  & \multicolumn{2}{c}{w/ CFG}    \\
Method                                               & Ratio & Type        & \#Tok. & \#Code   & Acc.      &           & Method                 &          &        & gFID$\downarrow$ & IS$\uparrow$ & gFID$\downarrow$ & IS$\uparrow$ \\ \midrule
\small{Taming-VQGAN}~{\tiny\cite{cvpr2021vqgan}}            & 16    & 2D VQ       & $16^2$   & $2^{10}$  & --        & 7.94      & Taming-Trans.          & 1.4B     & 256    & 15.78          & 78.3         & --             & --           \\
ViT-VQGAN~{\tiny\cite{ICLR2021VIT-VQGAN}}                   & 8     & 2D VQ       & $32^2$   & $2^{13}$  & 65.1      & 1.28      & VIM-L                  & 1.7B     & 1024   & 4.17           & 175.1        & --             & --           \\
LlamaGen~{\tiny\cite{NIPS2024LLaMAGen}}                     & 16    & 2D VQ       & $16^2$   & $2^{14}$  & 47.6      & 2.19      & LlamaGen-L             & 343M     & 256    & 3.80           & 248.3        & 3.07           & 256.1        \\
LlamaGen~{\tiny\cite{NIPS2024LLaMAGen}}                     & 16    & 2D VQ       & $16^2$   & $2^{14}$  & 47.6      & 2.19      & LlamaGen-XL            & 775M     & 256    & 3.39           & 227.1        & 2.62           & 244.1        \\
LlamaGen~{\tiny\cite{NIPS2024LLaMAGen}}                     & 16    & 2D VQ       & $16^2$   & $2^{14}$  & 47.6      & 2.19      & \small{LlamaGen-XXL}   & 1.4B     & 256    & --             & --           & 2.34           & 253.9        \\
\small{OmniTokenizer}~{\tiny\cite{wang2024Omnitokenizer}}   & 16    & \small{2D VQ+VAE}   & $16^2$   & $2^{13}$  & --        & 1.11      & GPT2                   & 650M     & 256    & 7.45           & 146.7        & --             & --           \\
VFMTok~{\tiny\cite{nips2025VFMTok}}                         & 16    & 2D VQ       & $16^2$   & $2^{14}$  & 69.4      & 0.89      & LlmaGen-B              & 111M     & 256    & 3.09           & 173.6        & 3.43           & 252.2        \\
VFMTok~{\tiny\cite{nips2025VFMTok}}                         & 16    & 2D VQ       & $16^2$   & $2^{14}$  & 69.4      & 0.89      & LlmaGen-L              & 343M     & 256    & 2.11           & 230.1        & 2.75           & 278.8        \\
UniTok$^\star$~{\tiny\cite{Ma2025UniTok}}                   & 16    & 2D MSQ      & $16^2$   & $2^{15}$  & \bf{70.8} & \bf{0.41} & \small{LlamaGen-XXL}   & 1.4B     & 256    & 2.51           & 216.7        & 2.77           & 227.5        \\ \midrule
\small{OpenMAGVIT2}~{\tiny\cite{luo2024Open-Magvit2}}       & 16    & 2D LFQ      & $16^2$   & $2^{18}$  & --        & 1.17      & LlamaGen-B             & 343M     & 256    & 3.08           & 258.3        & --             & --           \\
\small{OpenMAGVIT2}~{\tiny\cite{luo2024Open-Magvit2}}       & 16    & 2D LFQ      & $16^2$   & $2^{18}$  & --        & 1.17      & LlamaGen-XL            & 1.5B     & 256    & 2.33           & 271.8        & --             & --           \\
IBQ~{\tiny\cite{Shi2024IBQ}}                                & 16    & 2D LFQ      & $16^2$   & $2^{18}$  & --        & 1.00      & LlamaGen-B             & 342M     & 64     & 2.88           & 254.7        & --             & --           \\
IBQ~{\tiny\cite{Shi2024IBQ}}                                & 16    & 2D LFQ      & $16^2$   & $2^{18}$  & --        & 1.00      & LlamaGen-XL            & 1.1B     & 64     & \bf{2.14}      & \bf{279.0}   & --             & --           \\
FlowMo~{\tiny\cite{iccv2025FlowMo}}                         & 16    & \small{2D Diff.+LFQ} & $16^2$   & $2^{18}$  & --        & \bf{0.95} & LlamaGen-B             & 397M     & 256    & 4.30           & 274.0        & --             & --           \\
\midrule
Titok-S-128~{\tiny\cite{nips2024titok}}                     & 16    & 1D VQ       & 128      & $2^{12}$  & 46.6      & 1.71      & \small{MaskGIT-UViT-L} & 287M     & 8      & 4.61           & 166.7        & 2.50           & 278.7        \\
Titok-B-64~{\tiny\cite{nips2024titok}}                      & 16    & 1D VQ       & 64       & $2^{12}$  & 53.9      & 1.70      & MaskGIT-VIT            & 177M     & 8      & 3.08           & 192.5        & 2.48           & 214.7        \\
Titok-L-32~{\tiny\cite{nips2024titok}}                      & 16    & 1D VQ       & 32       & $2^{12}$  & 60.0      & 2.21      & MaskGIT-VIT            & 177M     & 8      & 3.15           & 173.0        & 2.77           & 199.8        \\
MergeVQ-GR~{\tiny\cite{cvpr2025MergeVQ}}                    & 16    & 1D LFQ      & 144      & $2^{18}$  & 77.9      & 1.48      & RandAR-L               & 343M     & 64     & --             & --           & 2.63           & 279.5        \\
MergeVQ-GR~{\tiny\cite{cvpr2025MergeVQ}}                    & 16    & 1D LFQ      & 256      & $2^{18}$  & 77.9      & 1.12      & MergeAR-L              & 343M     & 256    & 3.25           & 253.8        & --             & --           \\
GigaTok-SB~{\tiny\cite{iccv2025GigaTok}}                    & 16    & 1D VQ       & 256      & $2^{14}$  & 61.5      & 0.89      & LlamaGen-B             & 111M     & 256    & --             & --           & 3.83           & 233.3        \\
GigaTok-BL$^\star$~{\tiny\cite{iccv2025GigaTok}}            & 16    & 1D VQ       & 256      & $2^{14}$  & 64.1      & 0.51      & LlamaGen-B             & 111M     & 256    & --             & --           & 3.33           & 265.4        \\
GigaTok-BL~{\tiny\cite{iccv2025GigaTok}}                    & 16    & 1D VQ       & 256      & $2^{14}$  & 63.8      & 0.81      & \small{LlamaGen-XXL}   & 1.4B     & 256    & 2.03           & 238.5        & --             & --           \\
Hita~{\tiny\cite{iccv2025Hita}}                             & 16    & 1D VQ       & 569      & $2^{14}$  & 36.6      & 1.03      & LlamaGen-B             & 111M     & 569    & --             & --           & 4.33           & 238.9        \\
Hita~{\tiny\cite{iccv2025Hita}}                             & 16    & 1D VQ       & 569      & $2^{14}$  & 36.6      & 1.03      & LlamaGen-L             & 343M     & 569    & --             & --           & 2.86      & 267.3        \\
VAEVQ~{\tiny\cite{Yang2025VAEVQED}}                             & 16    & VAE+VQ       & 256      & $2^{14}$  & -      & 1.14      & LlamaGen-B             & 111M     & 256    & 4.68             & --           & --      & --        \\
\brow \bf{MergeTok-SB}                               & 16    & \small{1D VAE+VQ}   & 256      & $2^{14}$  & 73.8      & 0.97      & LlamaGen-B             & 111M     & 256    & 3.92             & 182.7           & 3.37           & 245.7        \\
\brow \bf{MergeTok-BL}$^\star$                       & 16    & \small{1D VAE+VQ}   & 256      & $2^{14}$  & 78.2      & \bf{0.48} & LlamaGen-B             & 111M     & 256    & 3.56           & 247.6        & 3.09           & 267.8   \\
\brow \bf{MergeTok-BL}                               & 16    & \small{1D VAE+VQ}   & 256      & $2^{14}$  & \bf{78.3} & 0.78      & \small{LlamaGen-XXL}   & 1.4B     & 256    & \bf{1.93}      & \bf{265.4}   & \bf{2.13}             & \bf{281.5}           \\
    \bottomrule
    \end{tabular}
    }
    \label{tab:in1k_gen_256_vq}
\end{table}

\paragraph{Generator setup.}
We separate the downstream generators by branch. \textit{(i) VQ branch -- discrete AR generation:} we train LlamaGen-B (111M, 12 blocks, 12 heads) and LlamaGen-XXL (1.4B, 48 blocks, 24 heads) under a WSD scheduler with base learning rate $1\times 10^{-4}$, decay ratio $0.2$, and a 1-epoch warm-up. AdaLN~\cite{iccv2023DiT} is not applied since it targets class-conditional setups. Batch sizes are $256$ for B/L and $512$ for XXL; all AR models are trained for 300 epochs on ImageNet-1K at $256{\times}256$. For CFG, we sweep the guidance scale with step $0.25$ and report the lowest gFID obtained. \textit{(ii) VAE branch -- continuous diffusion/flow generation:} we adopt DiT and SiT as denoising models on ImageNet-1K at $256{\times}256$, both with patch size $1$ and 1D absolute positional embeddings. The XL variants (675M) of DiT and SiT are trained for $3$M steps; lighter ablations use SiT-L trained for $400$K steps. Other hyperparameters follow each generator's official configuration.

\paragraph{Metrics.}
We report rFID for reconstruction fidelity, linear probing accuracy (Lin.~Acc) for representation quality, and gFID/IS (with and without CFG) for class-conditional generation. To diagnose VQ codebook health we additionally examine codebook usage, dead-code (Collapse) rate, and perplexity (Appendix~\ref{app:codebook_health}). Metric definitions and probing protocols follow standard practice and are summarized in Appendix~\ref{app:impl}.

\paragraph{Fairness controls.}
All main results are evaluated on ImageNet-1K at $256{\times}256$. We compare against baselines under matched token budgets: the tokenizer interface is fixed at 256 tokens, downstream generators are not modified to exploit merging, and comparisons reuse the same generator family whenever available (\textit{e.g.}, MergeTok-BL vs.\ GigaTok-BL both with LlamaGen-XXL). For small gaps we use cautious wording (\textit{e.g.}, ``slightly improves over''), and we do not claim improvements over baselines whose numbers we do not strictly dominate.

\begin{table}[t]
    \centering
    \caption{\textbf{System-level comparison of continuous tokenizers on ImageNet 256$\times$256}. We report rFID for reconstruction, and gFID/IS for class-conditional generation. ``CFG": classifier-free guidance.
    }
    \setlength{\tabcolsep}{1.5pt}
    \resizebox{1.0\linewidth}{!}{
    \footnotesize
    \begin{tabular}{l|cccc|ccc|cc|cc}
    \toprule
Tokenizer                                 & Type              & \#Tok. & Down. & Tok.             & Generator                 & Training & \#Param. & \multicolumn{2}{c|}{w/o CFG}    & \multicolumn{2}{c}{w/ CFG}      \\
Method                                    &                   &          & ratio & rFID$\downarrow$ & Method                    & Epochs   &          & gFID$\downarrow$ & IS$\uparrow$ & gFID$\downarrow$ & IS$\uparrow$ \\ \midrule
TiTok-BL-KL~{\tiny\cite{nips2024titok}}          & 1D VAE            & 64       & 16    & 1.25             & SiT-L/1                   & 400      & 458M     & 23.35            & 54.7         & --               & --           \\
MAR~{\tiny\cite{NIPS2024MAR}}                    & 2D VQ             & 256      & 16    & 1.22             & MAR-H/1                   & 800      & 943M     & 2.35             & 227.8        & 1.55             & 303.7        \\
SoftVQ-S$^\star${\tiny\cite{cvpr2025SoftVQVAE}}  & 1D VQ             & 256      & 16    & 0.80             & SiT-L/1                   & 400      & 458M     & 9.21             & 93.6         & --               & --           \\
SoftVQ-BL$^\star${\tiny\cite{cvpr2025SoftVQVAE}} & 1D VQ             & 64       & 16    & 0.65             & DiT-XL/1                  & 300      & 675M     & 6.53             & 131.9        & 3.11             & 268.3        \\
SoftVQ-BL$^\star${\tiny\cite{cvpr2025SoftVQVAE}} & 1D VQ             & 64       & 16    & 0.65             & SiT-XL/1                  & 300      & 675M     & 5.80             & 143.5        & 1.88             & 287.9        \\
SoftVQ-L$^\star${\tiny\cite{cvpr2025SoftVQVAE}}  & 1D VQ             & 64       & 16    & 0.61             & SiT-XL/1                  & 300      & 675M     & 5.35             & 151.2        & 1.86             & 293.6        \\
\midrule
SD-VAE~{\tiny\cite{iclr2013VAE}}                 & 2D VAE            & $32^2$   & 8     & 0.61             & DiT-XL/2                  & 1400     & 675M     & 9.62             & 121.5        & 2.27             & 278.2        \\
SD-VAE~{\tiny\cite{iclr2013VAE}}                 & 2D VAE            & $32^2$   & 8     & 0.61             & SiT-XL/2                  & 1400     & 675M     & --               & --           & 2.62             & 252.2        \\
REPA~{\tiny\cite{Yu2024REPA}}                    & 2D VAE            & $32^2$   & 8     & 0.61             & DiT-XL/2                  & 800      & 675M     & 5.78             & 158.3        & \bf{1.29}        & 306.3        \\
DC-AE-f32~{\tiny\cite{iclr2025DCAE}}             & 2D VAE            & $8^2$    & 32    & 0.69             & DiT-XL/1                  & 500      & 675M     & 9.56             & --           & 2.84             & --           \\
DC-AE-f32~{\tiny\cite{iclr2025DCAE}}             & 2D VAE            & $8^2$    & 32    & 0.69             & SiT-XL/1                  & 500      & 675M     & 7.47             & --           & 2.41             & --           \\
VAVAE~{\tiny\cite{cvpr2025VAVAE}}                & 2D VAE            & $16^2$   & 16    & 0.61             & \footnotesize{LightningDiT-XL/1} & 800      & 675M     & 2.17             & 205.6        & 1.35             & 295.3        \\
RAE (\small{DINOv2-S})~{\tiny\cite{Zheng2025RAE}}        & 1D VAE            & 256      & 16    & \bf{0.49}        & DiT-XL/1                  & 800      & 675M     & \bf{1.87}        & \bf{209.7}   & 1.41             & \bf{309.4}   \\
\brow \bf{MergeTok-BL}                    & \small{1D VAE+VQ} & 256      & 16    & \bf{0.47}        & DiT-XL/1                  & 800      & 675M     & 1.91             & 211.4        & 1.44             & 304.1        \\
\brow \bf{MergeTok-BL}                    & \small{1D VAE+VQ} & 256      & 16    & \bf{0.47}        & SiT-XL/1                  & 800      & 675M     & \bf{1.79}        & \bf{217.6}   & \bf{1.29}        & \bf{311.7}   \\ 
    \bottomrule
    \end{tabular}
    }
    \label{tab:in1k_gen_256_vae}
\end{table}

\subsection{Main Results on ImageNet-256}
\label{sec:exp_main}

\paragraph{Discrete tokenization: reconstruction.}
Under the same 256-token interface, MergeTok-BL achieves $0.78$ rFID, slightly improving over GigaTok-BL at $0.81$ and outperforming MergeVQ-GR at $1.12$ (Table~\ref{tab:in1k_gen_256_vq}). The SB variant reaches $0.97$ rFID at a much smaller capacity. With the frozen DINO discriminator, MergeTok-BL$^{\star}$ reaches $0.48$ rFID, marginally improving over GigaTok-BL$^{\star}$ at $0.51$ under the same training protocol. UniTok$^{\star}$ achieves a lower $0.41$ rFID but relies on multi-codebook quantization, while MergeTok keeps a single $16{,}384$-entry codebook; we therefore report this as a competitive result rather than a universal claim of reconstruction superiority.

\paragraph{Discrete tokenization: representation.}
Reconstruction alone cannot reveal whether the tokenizer organizes semantics. Linear probing directly evaluates the discriminative structure of the learned token space (Table~\ref{tab:in1k_gen_256_vq}). MergeTok-BL attains $78.3\%$ Lin.~Acc, the highest among the listed discrete tokenizers, exceeding VFMTok ($69.4\%$), UniTok$^{\star}$ ($70.8\%$), GigaTok-BL ($63.8\%$), and the concurrent merge-based MergeVQ-GR ($77.9\%$). The SB variant improves over GigaTok-SB ($61.5\%$) by $12.3$ points, providing evidence that the ToMe-derived source map and merged-token alignment contribute to semantic organization rather than only pixel-level reconstruction.

\paragraph{Discrete tokenization: AR generation.}
We test whether the discrete tokens produced by the VQ branch remain compatible with strong autoregressive generators. With LlamaGen-XXL, MergeTok-BL obtains $1.93$ gFID without CFG and $2.13$ with CFG, with IS of $265.4$ and $281.5$ respectively, surpassing GigaTok-BL paired with the same LlamaGen-XXL ($2.03$ gFID without CFG). The smaller SB$+$LlamaGen-B configuration delivers $3.92$/$3.37$ gFID without/with CFG. These numbers support the hypothesis that the VQ branch is generator-friendly and does not require any merge-aware modification on the generator side.

\paragraph{Continuous tokenization: reconstruction and diffusion/flow generation.}
On the continuous side (Table~\ref{tab:in1k_gen_256_vae}), MergeTok-BL achieves $0.47$ rFID, marginally improving over RAE at $0.49$ and outperforming SoftVQ-L ($0.61$), DC-AE-f32 ($0.69$), VAVAE ($0.61$), and SD-VAE ($0.61$). When paired with diffusion/flow generators, MergeTok-BL yields $1.91$ gFID with DiT-XL/1 and $1.79$ with SiT-XL/1; with CFG, SiT-XL/1 reaches $1.29$ gFID, matching the best entry in the table. These results indicate that the VAE branch is not merely a reconstruction pathway but also provides a continuous token interface suitable for modern diffusion and flow generators.

\paragraph{Takeaway.}
Tables~\ref{tab:in1k_gen_256_vq} and~\ref{tab:in1k_gen_256_vae} jointly support the claim that MergeTok is not just a high-rFID tokenizer: it provides a unified 256-token interface that simultaneously supports discrete AR generation, continuous diffusion/flow generation, and semantically structured representation learning under matched token budgets and matched generators.

\paragraph{Extended evaluations in the appendix.}
Beyond the ImageNet-256 comparisons in Tables~\ref{tab:in1k_gen_256_vq} and~\ref{tab:in1k_gen_256_vae}, Appendix~\ref{app:viz} provides qualitative reconstruction and generation results for both branches, and Appendix~\ref{app:exp} extends the evaluation to higher-resolution reconstruction at $512{\times}512$ and $1024{\times}1024$, class-conditional generation at $512{\times}512$, and a cross-domain MS-COCO reconstruction setting. These extended results complement the main comparisons by showing that the observed advantages are not limited to the default ImageNet-256 setting.

\begin{table}[t]
\caption{\textbf{Component-level ablations on MergeTok-SB.} Each sub-table targets one design question raised in Sec.~\ref{sec3.2_framework}--Sec.~\ref{Sec3.4}. All entries report VAE-rFID$\downarrow$\,/\,VQ-rFID$\downarrow$ on ImageNet $256{\times}256$; the bottom row (\textbf{bold}) is the default MergeTok-SB configuration. \textbf{(a)}~Progressive integration of the dual VAE/VQ branch, the ToMe-derived source map, and the merged-token alignment loss. \textbf{(b)}~Loss objectives, where $\mathcal{L}_{\mathrm{group}}=\mathcal{L}_{\mathrm{div}}+\mathcal{L}_{\mathrm{cons}}$ (Eqs.~\ref{eq:loss_div}--\ref{eq:loss_cons}) and $\mathcal{L}_{\mathrm{ent}}$ denotes the GigaTok-style entropy regularizer. \textbf{(c)}~Granularity at which the alignment loss $\mathcal{L}_{\mathrm{align}}$ is applied.}
\label{tab:ablation_branch}
\vspace{-2pt}
\centering
\footnotesize
\setlength{\tabcolsep}{2pt}
\begin{minipage}[t]{0.34\textwidth}
\centering
\subcaption{Progressive component integration.}
\label{tab:ablation_branch_a}
\begin{tabular}{lrr}
\toprule
Configuration & VAE & VQ \\ \midrule
GigaTok-SB (w/o Dist) & -- & 1.12 \\
VAE only & 0.81 & -- \\
Dual-branch (VQ+VAE) & 0.83 & 1.06 \\
~~$+$ ToMe-derived src.\ map & 0.67 & 1.01 \\
~~$+$ Merged-token align. & 0.62 & 0.97 \\
\rowcolor[HTML]{CFEFFF}
\textbf{MergeTok-SB (full)} & \textbf{0.59} & \textbf{0.96} \\
\bottomrule
\end{tabular}
\end{minipage}%
\hfill
\begin{minipage}[t]{0.34\textwidth}
\centering
\subcaption{Loss component combinations.}
\label{tab:ablation_loss}
\renewcommand{\arraystretch}{1.15}
\begin{tabular}{lrr}
\toprule
Loss configuration & VAE & VQ \\ \midrule
$\mathcal{L}_{\mathrm{group}}$ only & 0.78 & 1.01 \\
$\mathcal{L}_{\mathrm{align}}$ only & 0.65 & 1.08 \\
$\mathcal{L}_{\mathrm{ent}}$ only & 0.77 & 1.09 \\
$\mathcal{L}_{\mathrm{group}}{+}\mathcal{L}_{\mathrm{align}}{+}\mathcal{L}_{\mathrm{ent}}$ & 0.62 & 1.02 \\
\rowcolor[HTML]{CFEFFF}
$\bm{\mathcal{L}_{\mathrm{group}}{+}\mathcal{L}_{\mathrm{align}}}$ \textbf{(Ours)} & \textbf{0.59} & \textbf{0.97} \\
\bottomrule \\
\end{tabular}
\renewcommand{\arraystretch}{1.0}
\end{minipage}%
\hfill
\begin{minipage}[t]{0.30\textwidth}
\centering
\subcaption{Alignment granularity.}
\label{tab:ablation_alignment_granularity}
\renewcommand{\arraystretch}{1.3}
\begin{tabular}{lrr}
\toprule
Granularity & VAE & VQ \\ \midrule
No alignment & 0.67 & 1.01 \\
{[CLS]} only ($1$) & 0.64 & 1.00 \\
All patches ($256$) & 0.62 & 0.98 \\
\rowcolor[HTML]{CFEFFF}
\textbf{Merged (128, Ours)} & \textbf{0.59} & \textbf{0.96} \\
\bottomrule
\end{tabular}
\renewcommand{\arraystretch}{1.0}
\end{minipage}
\end{table}

\subsection{Ablation Study}
\label{sec:ablation}

Table~\ref{tab:ablation_branch} reports the main design-oriented ablations on MergeTok-SB, and Fig.~\ref{fig:token_vs_fid} analyzes the effect of token granularity at $K^{*}\!\in\![64, 256]$ with the minimum near $K^{*}{=}128$, which we adopt in all main results. Appendix~\ref{app:ablation} further expands these ablations from four complementary perspectives: Appendix~\ref{app:ablation_component} isolates the VQ branch, the VAE branch, and the full joint training; Appendix~\ref{app:codebook_health} reports codebook-health diagnostics including usage, collapse rate, perplexity, active codes, dead codes, and entropy; Appendix~\ref{app:multi_metric} evaluates the same variants across reconstruction, representation, and generation metrics; and Appendix~\ref{app:alt_grouping} replaces the ToMe-derived source map with alternative grouping strategies. Together, the main-paper and appendix ablations verify both the component-level necessity and the multi-metric robustness of MergeTok.

\paragraph{Component integration.}
Table~\ref{tab:ablation_branch_a} follows the construction of MergeTok step by step. Starting from the GigaTok-SB VQ baseline ($1.12$ VQ-rFID), simply adding a parallel VAE branch reduces VQ-rFID only marginally to $1.06$ and yields VAE-rFID $0.83$, suggesting that joint training alone is insufficient. Introducing the ToMe-derived source map drops VAE-rFID to $0.67$ and VQ-rFID to $1.01$, indicating that content-adaptive grouping provides an effective semantic bottleneck shared by both branches. Adding the merged-token alignment loss further improves both metrics to $0.62$/$0.97$, and the full model reaches $0.59$/$0.96$. Note that the VQ branch's forward path bypasses ToMe; the source map enters only as a structural signal that drives the group-aware losses. The fact that the VQ branch still improves under joint training is consistent with continuous gradients from the VAE branch flowing through the shared encoder.

\paragraph{Loss objectives.}
Table~\ref{tab:ablation_loss} separates the roles of the merge-aware objectives. Among single-loss variants, $\mathcal{L}_{\mathrm{align}}$ gives the best VAE result ($0.65$) while $\mathcal{L}_{\mathrm{group}}=\mathcal{L}_{\mathrm{div}}+\mathcal{L}_{\mathrm{cons}}$ gives the best VQ result ($1.01$), supporting the view that the two losses regularize complementary aspects of the unified tokenizer. Their combination achieves the best overall trade-off ($0.59$/$0.97$). Adding the GigaTok-style entropy regularizer $\mathcal{L}_{\mathrm{ent}}$ on top slightly worsens both branches ($0.62$/$1.02$), so we omit it from the default objective.

\paragraph{Further ablations in the appendix.}
Beyond Table~\ref{tab:ablation_branch}, we defer three additional ablation studies to the appendix: codebook-health diagnostics that evaluate MergeTok-SB on usage, active/dead codes, collapse rate, perplexity, and entropy (Appendix~\ref{app:codebook_health}); a multi-metric ablation jointly reporting rFID, PSNR, SSIM, linear probing accuracy, and AR/diffusion gFID (Appendix~\ref{app:multi_metric}); and a comparison against alternative grouping strategies ($k$-means, spatial grid, random, and no grouping) under matched group count $K{=}128$ (Appendix~\ref{app:alt_grouping}). These studies extend the rFID-based conclusions in Table~\ref{tab:ablation_branch} to codebook stability, semantic representation, downstream generation, and the specificity of the ToMe-derived source map.

\section{Conclusion}

We introduce \emph{MergeTok}, a unified visual tokenizer bridging continuous (VAE) and discrete (VQ) modeling via token merging. The merging provides a semantic conduit, enabling (i) merged-token alignment to regularize the VAE latent space; and (ii) group-aware quantization to stabilize VQ training and improve codebook utilization. A shared encoder-decoder is trained end-to-end with a single joint objective and light merge-ratio sampling.
On ImageNet-1K at \(256\times256 \), \emph{MergeTok} substantially improves rFID over continuous and discrete baselines, while producing 256-token sequences with strong semantic organization, suitable for both AR and diffusion generators. Ablations further show that the ToMe-derived grouping outperforms alternative clustering strategies (Appendix~\ref{app:alt_grouping}). This shows that a unified architecture can be simultaneously semantics-aware and generator-friendly.


{
\small
\bibliographystyle{plainnat}
\bibliography{reference}
}

\newpage
\appendix
\renewcommand\thefigure{A\arabic{figure}}
\renewcommand\thetable{A\arabic{table}}
\setcounter{table}{0}
\setcounter{figure}{0}
\setcounter{page}{1}

\clearpage
\appendix
\begin{center}
{\LARGE\bfseries Appendix for MergeTok}
\end{center}
\vspace{0.8em}

\section*{Roadmap}
The appendix is organized to follow the structure of the main paper, with each part providing deeper coverage of the corresponding main-text section. %

\smallskip
\noindent\textbf{Part~I -- Extended Discussion} (App.~\ref{app:related}--\ref{app:recent_works}, cited in §\ref{sec:related_work}) provides a structured design comparison with concurrent hybrid tokenizers (App.~\ref{app:related}) and discussion of recently proposed concurrent works (App.~\ref{app:recent_works}), supplementing the Related Work section. %

\smallskip
\noindent\textbf{Part~II -- Technical Details} (App.~\ref{app:tome}--\ref{app:unmerge}, cited in §\ref{sec3.2_framework}) derives the token merging mechanism in encoding (App.~\ref{app:tome}) and the source map unmerging process with a worked example (App.~\ref{app:unmerge}), establishing how $S$ couples the continuous and discrete branches, supplementing the Methods section. %

\smallskip
\noindent\textbf{Part~III -- Implementation Details} (App.~\ref{app:impl}) records full architectural specifications, training hyperparameters, and evaluation protocols for the tokenizer and downstream generators, supporting the experimental setup of §\ref{sec:exp}. %

\smallskip
\noindent\textbf{Part~IV -- Extended Experimental Results} (App.~\ref{app:viz}--\ref{app:exp}, cited in §\ref{sec:ablation} and §\ref{sec:exp_main}) presents additional visualizations (App.~\ref{app:viz}), more ablation studies (App.~\ref{app:ablation}), and more results on benchmarks (App.~\ref{app:exp}), expanding the experimental results of §\ref{sec:exp}. %

\clearpage
\begin{center}
\rule{0.85\linewidth}{0.4pt}\\[0.4em]
{\large\bfseries Part~I\;\;|\;\;Extended Discussion}\\[0.25em]
{\itshape\small Design comparison with concurrent hybrid tokenizers and discussion of recently proposed concurrent works.}\\[0.3em]
\rule{0.85\linewidth}{0.4pt}
\end{center}
\vspace{0.4em}

This part supplements the related-work discussion in \S\ref{sec:related_work} by positioning MergeTok with respect to concurrent and closely related tokenizer designs.
Appendix~\ref{app:related} provides a structured comparison with hybrid or unified tokenizers along five axes -- branch composition, role of token merging, generator compatibility, and VQ regularization.
Appendix~\ref{app:recent_works} discusses recently proposed concurrent works and clarifies how MergeTok differs in using the ToMe-derived source map $S$ as an explicit bridge between the continuous and discrete branches.

\section{Design Comparison with Concurrent Hybrid Tokenizers}
\label{app:related}

Table~\ref{tab:design_comparison} compares MergeTok with three concurrent works along five design axes: MergeVQ~\cite{cvpr2025MergeVQ}, Wave-Particle~\cite{Chen2025DCC-VT}, and UniTok~\cite{Ma2025UniTok}.

\begin{table}[h]
\centering
\caption{\textbf{Design comparison with concurrent hybrid and unified tokenizers.}}
\label{tab:design_comparison}
\small
\setlength{\tabcolsep}{5pt}
\resizebox{1.0\linewidth}{!}{
\begin{tabular}{l|cccc}
\toprule
\textbf{Design Axis}      & \textbf{MergeVQ} & \textbf{Wave-Particle} & \textbf{UniTok} & \textbf{MergeTok (Ours)} \\ \hline
Continuous Branch         & \textemdash      & KL-AE                  & \textemdash     & KL-AE                    \\
Discrete Branch           & LFQ              & VQ                     & VQ              & VQ                       \\
Role of Token Merging     & Token Compression& \textemdash            & \textemdash     & Semantic Bridge          \\
Generator                 & AR               & AR                     & AR              & AR + Diff                \\
VQ Regularization         & Entropy          & \textemdash            & \textemdash     & Group Constraint Loss    \\
\bottomrule
\end{tabular}
}
\end{table}

Three differences are worth noting. First, MergeVQ uses ToMe as a compression mechanism, producing variable-length quantized sequences, whereas MergeTok uses ToMe as a structural bridge: the source map $S$ couples the continuous and discrete branches via alignment (Eq.~\ref{eq:loss_align}) and group-aware regularization (Eqs.~\ref{eq:loss_div},~\ref{eq:loss_cons}), and the tokenizer always outputs a fixed 256-token sequence. Second, Wave-Particle builds two parallel branches with partial sharing but without a source-map interface, while MergeTok transfers $S$ explicitly from the VAE branch to the VQ branch through a fully shared encoder and decoder. Third, UniTok targets AR generation only, whereas MergeTok simultaneously supports AR generation via the VQ branch and diffusion generation via the VAE branch within a single tokenizer. Codebook health diagnostics in Table~\ref{tab:codebook_health} further confirm that MergeTok's group constraint loss achieves high codebook utilization without enlarging codebook capacity.

\section{Discussion of Recently Proposed Concurrent Works}
\label{app:recent_works}

Several recent or concurrent works share goals or mechanisms with MergeTok and merit explicit comparison.

\textbf{CODA}~\cite{Liu2025CODA} repurposes pretrained continuous VAEs into discrete tokenizers by applying post-hoc quantization rather than joint training. MergeTok differs in that both branches are trained jointly from scratch with a shared encoder, and the ToMe-derived source map $S$ serves as an architectural bridge between the two branches.

\textbf{MGVQ}~\cite{Liang2024MGVQ} improves VQ tokenizers via multi-group quantization, decomposing a large codebook into coordinated subgroups to reduce collapse. Although the naming is similar, MGVQ's grouping operates on the codebook side by decomposing the codebook into subspaces, whereas MergeTok's grouping operates on the token side by using $S$ to partition tokens into semantic groups for regularization. The two approaches are complementary and could in principle be combined.

\textbf{MingTok}~\cite{Zhao2024MingTok} proposes a continuous unified tokenizer for autoregressive understanding and generation. It shares MergeTok's goal of unification but relies solely on continuous representations. MergeTok's discrete branch and ToMe-bridged design provide an alternative path that retains compatibility with discrete AR generators such as LlamaGen without requiring continuous-only AR architectures.

\textbf{UniToken}~\cite{Xu2024UniToken} combines discrete and continuous representations in a unified visual encoding for multimodal downstream tasks. In contrast, MergeTok's discrete and continuous branches share the encoder and decoder and are coupled via $S$, whereas UniToken concatenates separately trained representations for downstream consumption.

\textbf{VTBench}~\cite{Gao2024VTBench} provides a standardized benchmark for visual tokenizers in autoregressive image generation. Evaluation under the VTBench protocol would strengthen external validity; we plan to include such evaluation in a future revision.

\clearpage
\begin{center}
\rule{0.85\linewidth}{0.4pt}\\[0.4em]
{\large\bfseries Part~II\;\;|\;\;Technical Details}\\[0.25em]
{\itshape\small The token merging and unmerging mechanism that underpins MergeTok's shared-encoder design.}\\[0.3em]
\rule{0.85\linewidth}{0.4pt}
\end{center}
\vspace{0.4em}

This part provides the technical details that support the method description in \S\ref{sec3.2_framework}.
Appendix~\ref{app:tome} expands the token merging process used inside the VAE branch of the attention encoder, including the per-layer schedule and the $\log s$ size-aware attention correction.
Appendix~\ref{app:unmerge} works through a concrete source-map and unmerging example, showing how $S$ and its one-hot matrix form $A$ preserve token-to-group correspondence.
Together, these details explain how the source map serves as the structural interface that couples the VAE and VQ branches.

\section{Token Merging in Encoding}
\label{app:tome}

Following ToMe~\cite{iclr2022ToMe}, MergeTok inserts a lightweight merge unit into each Transformer encoder block to progressively reduce the number of spatial tokens while preserving semantic content. Given the patch-token sequence \(Z_L \in \mathbb{R}^{L \times D}\), ToMe performs merging in four steps:
\begin{enumerate}
    \item Evenly partition tokens into two sets \(A\) and \(B\) (e.g., by odd/even indices).
    \item For each token in \(A\), find its most similar token in \(B\) according to the current-layer attention features.
    \item Select the top pairs for merging under a pre-defined schedule.
    \item Aggregate features within each selected pair (e.g., by averaging) to form merged tokens.
\end{enumerate}

After one merge operation, the sequence length is reduced from \(L\) to \(K\) (\(K<L\)), producing a compressed sequence \(Z_K \in \mathbb{R}^{K \times D}\) that is fed into the next layer.
Token similarity is computed using attention features from the current layer, typically the self-attention keys \( \mathrm{Key} \)  to match tokens across \(A\) and \(B\).
Since each merged token may represent multiple original tokens, ToMe tracks the \emph{token size} \(s\) (the number of originals aggregated into a surviving token) and compensates its influence in self-attention by adding \(\log s\) to the attention logits:
\begin{equation}
\mathbf{A} = \mathrm{softmax}\!\left(\frac{Q\,\mathrm{Key}^{\top}}{\sqrt{d}} + \log s\right),
\end{equation}
where \(Q\) and \(\mathrm{Key}\) denote the query and key matrices, \(d\) is the attention head dimension, and \(s\) is broadcast to match the logits shape.
Across layers, ToMe follows a merging schedule that controls how many pairs are merged, trading off efficiency and fidelity. In practice, we adopt a decreasing schedule (e.g., square decreasing) where the number of merges increases with depth to encourage stronger semantic abstraction at higher layers.

\section{Source Map for Token Unmerging}
\label{app:unmerge}

In this section, we provide a concrete example to illustrate how token merging and the subsequent unmerging process can be expressed using the \emph{source map} $S$ and its one-hot matrix form $A$.

As introduced in Sec.~\ref{sec3.2_framework}, the VAE branch performs token merging during encoding: it compresses the original length-$L$ token sequence $Z_L \in \mathbb{R}^{L \times D}$ into a compact sequence $Z_K^{(\mathrm{vae})} \in \mathbb{R}^{K \times D}$, accompanied by a discrete source map $S$ (cf.~Eq.~\ref{equ:vae_forward}). During decoding, the reconstruction process conditions on both $Z_K^{(\mathrm{vae})}$ and $S$ to recover the original structure. Each entry $S_i$ (for $i = 1,\dots,L$) indicates the index of the merged token to which the $i$-th original token belongs.

\paragraph{Example.} Consider an example with $L = 5$ and $K = 3$. Let the original token sequence be denoted as $Z_L = [z_1, z_2, z_3, z_4, z_5]^\top \in \mathbb{R}^{5 \times D}$, where each $z_i \in \mathbb{R}^D$ is a token embedding.
Suppose the source map is given by:
\begin{equation}
S = [0, 0, 1, 1, 2],
\end{equation}
indicating that the first two tokens are assigned to cluster 0, the next two to cluster 1, and the last token to cluster 2.

This assignment can be encoded by a one-hot \emph{source matrix} $A \in \{0,1\}^{L \times K}$, where each row $i$ has a 1 in column $S_i$:
\begin{equation}
A_{ij} =
\begin{cases}
1, & \text{if } S_i = j \\
0, & \text{otherwise}
\end{cases}
\end{equation}

For the given $S$, the source matrix becomes:
\begin{equation}
A =
\begin{pmatrix}
1 & 0 & 0 \\
1 & 0 & 0 \\
0 & 1 & 0 \\
0 & 1 & 0 \\
0 & 0 & 1
\end{pmatrix}
\in \{0,1\}^{5 \times 3}
\end{equation}

\paragraph{Token merging.} Given the original sequence $Z_L$, the compressed sequence of merged tokens $Z_K = [z^{(\mathrm{m})}_0, z^{(\mathrm{m})}_1, z^{(\mathrm{m})}_2]^\top \in \mathbb{R}^{K \times D}$ can be computed by:
\begin{equation}
Z_K = A^\top Z_L
\end{equation}

In this case:
\begin{equation}
z^{(\mathrm{m})}_0 = \frac{z_1 + z_2}{2}, \quad
z^{(\mathrm{m})}_1 = \frac{z_3 + z_4}{2}, \quad
z^{(\mathrm{m})}_2 = z_5
\end{equation}

\paragraph{Unmerging.} The reconstruction restores the token sequence layout by broadcasting each merged token back to its original positions:
\begin{equation}
\hat{Z}_L = A Z_K.
\end{equation}

For this example:
\begin{equation}
\hat{Z}_L =
A Z_K =
\begin{pmatrix}
1 & 0 & 0 \\
1 & 0 & 0 \\
0 & 1 & 0 \\
0 & 1 & 0 \\
0 & 0 & 1
\end{pmatrix}
\begin{pmatrix}
(z^{(\mathrm{m})}_0)^\top \\
(z^{(\mathrm{m})}_1)^\top \\
(z^{(\mathrm{m})}_2)^\top
\end{pmatrix}
=
\begin{pmatrix}
(z^{(\mathrm{m})}_0)^\top \\
(z^{(\mathrm{m})}_0)^\top \\
(z^{(\mathrm{m})}_1)^\top \\
(z^{(\mathrm{m})}_1)^\top \\
(z^{(\mathrm{m})}_2)^\top
\end{pmatrix}
\end{equation}

This demonstrates how the source map $S$ and its matrix form $A$ can be used to implement both token merging and unmerging using simple matrix multiplication.
Beyond efficient aggregation and broadcasting, $S$ and $A$ retain the spatial correspondence between original and merged tokens. Each entry in $S$ records the assignment of an original token to its semantic group, encoding its relative spatial location within the compressed representation. This spatial mapping enables a hybrid decoder architecture. After the semantic tokens $Z_K^{(\mathrm{vae})}$ are decoded by a transformer decoder, the source map guides the unmerging process, distributing semantic content back to spatial positions in the form of $\hat{Z}_L$. A pixel-level decoder can then further refine $\hat{Z}_L$ to reconstruct the image. The source map thus serves as a bridge between semantic representations and fine-grained spatial reconstructions.

\clearpage
\begin{center}
\rule{0.85\linewidth}{0.4pt}\\[0.4em]
{\large\bfseries Part~III\;\;|\;\;Implementation Details}\\[0.25em]
{\itshape\small Architectural specifications, training hyperparameters, and evaluation protocols for full reproducibility.}\\[0.3em]
\rule{0.85\linewidth}{0.4pt}
\end{center}
\vspace{0.4em}

\begin{table}[!ht]
    \centering
    \caption{
    Implementation details and configuration of network architecture, hyperparameters of loss functions, and training settings for the two versions of MergeTok tokenizers on ImageNet-1K.
    }
    \setlength{\tabcolsep}{6pt}
    \small
\begin{tabular}{l|cc}
    \toprule
Settings                    & MegreTok-SB                  & MergeTok-BL                                 \\ \midrule
Channels               & 256                 & 256                                \\
CNN Stage number            & 5                  & 5                                 \\
Channel multiplier          & [1, 1, 2, 2, 4]       & [1, 1, 2, 2, 4]                   \\
Encoder Attention Blocks            & 6      & 12                 \\
Encoder Attention Heads          & 8 & 12       \\
Encoder Attention Dim             &    512                & 768                              \\
Decoder Attention Blocks             &      12              & 24                              \\
Decoder Attention Heads             &       12             & 16                             \\
Decoder Attention Dim             &     768               & 1024                              \\
Vocabulary size             &       16384             & 16384                              \\ \midrule
Discriminator loss          &     0.5               & 0.5                                   \\
Perceptual loss             &        1.0            & 1.0                                   \\
VQ rec loss &         1.0           & 1.0                                  \\
VAE rec loss &       1.0             & 1.0                                  \\
Diversity loss           &       0.05             & 0.05                                   \\
Consistency loss           &   0.05                 & 0.05                                   \\
Commitment loss             &    0.25                & 0.25                                  \\
Alignement loss             & 0.5                & 0.5                               \\ \midrule
Optimizer                   &    AdamW                & AdamW                                 \\
($\beta_1$, $\beta_2$)      &    (0.9, 0.95)                & (0.9, 0.95)                            \\
Weight decay                &    1e-4                & 0.0                                   \\
Training epochs             & 200                & 200                               \\
Base learning rate          &  1e-4                  & 1e-4                                  \\
Batch size                  &     256               & 256                                   \\
LR scheduler                &    cosine\_v2            & cosine\_v2                           \\
 \midrule
\#Param. of Attn Encoder         & 19M              & 86M                           \\
\#Param. of Attn Decoder         & 86M              & 329M                           \\

    \bottomrule
    \end{tabular}
    \label{tab:mergetok_config}
\end{table}

\begin{table}[!ht]
    \centering
    \caption{
    Configuration of discrete and continuous visual generators with MergeTok for image generation on ImageNet-1K.
    }
    \setlength{\tabcolsep}{5pt}
    \resizebox{0.99\linewidth}{!}{
    \small
\begin{tabular}{l|ccccc}
    \toprule
Settings               & LlamaGen-B        & LlamaGen-L        & LLamaGen-XXL      & DiT-XL              & SiT-XL              \\ \midrule
Attention heads        & 12                & 24                & 48                & 16                  & 16                  \\
Input dim.             & 768               & 1024              & 1536              & \multicolumn{2}{c}{$16\times 16\times 4$} \\
Num. layers            & 12                & 24                & 1.4B              & 28                  & 28                  \\
Dropout                & 0.1               & 0.1               & 0.1               & 0                   & 0                   \\
Mask schedule          & Arccos            & Arccos            & Arccos            & --                  & --                  \\
Label smoothing        & 0.1               & 0.1               & 0.1               & --                  & --                  \\
Sampler                & --                & --                & --                & \multicolumn{2}{c}{Euler-Maruyama}        \\
Gen. steps             & 256               & 256               & 256               & 250                 & 250                 \\
\# Parameter           & 111M              & 343M              & 1.4B              & 675M                & 675M                \\ \midrule
Optimizer              & \multicolumn{3}{c}{AdamW}                                 & \multicolumn{2}{c}{AdamW}                 \\
($\beta_1$, $\beta_2$) & \multicolumn{3}{c}{(0.9, 0.95)}                           & \multicolumn{2}{c}{(0.9, 0.999)}          \\
Weight decay           & 5e-2              & 5e-2              & 5e-2              & 0                   & 0                   \\
Training epochs        & 300               & 300               & 300               & 400                 & 400                 \\
Base learning rate     & $1\times 10^{-4}$ & $1\times 10^{-4}$ & $1\times 10^{-4}$ & $1\times 10^{-4}$   & $1\times 10^{-4}$   \\
Batch size             & 256               & 256               & 512               & 256                 & 256                 \\
LR scheduler           & WSD               & WSD               & WSD               & --                  & --                  \\
    \bottomrule
    \end{tabular}
    }
    \label{tab:visual_generator}
\end{table}

\section{Implementation Details}
\label{app:impl}
\paragraph{MergeTok Models.}
We provide two parameter scales for the MergeTok tokenizer. Based on GigaTok~\cite{iccv2025GigaTok}, the SB configuration employs a small-size encoder coupled with a base-size decoder, while the BL configuration uses a base-size encoder and a large-size decoder. The detailed architectural specifications and training hyperparameters of both variants are reported in Table~\ref{tab:mergetok_config}.
For the autoregressive generator, we adopt Llama-Gen as the backbone architecture \cite{NIPS2024LLaMAGen} and instantiate three model sizes, namely Base, Large, and XX-Large (XXL). Their parameter counts and experimental settings are summarized in Table~\ref{tab:visual_generator}.
For the diffusion generator, we adopt SiT and DiT as the backbone~\cite{iccv2023DiT, eccv2024SIT}. Our DiT and SiT generators closely follow the original architectural designs. For DiT, we adopt the standard Transformer-based diffusion backbone operating on 1D latent token sequences~\cite{cvpr2022ldm}, using 1D absolute positional encodings~\cite{iclr2021ViT} and keeping the depth, embedding dimension, number of attention heads, and MLP width identical to the corresponding DiT variants in the original work. Similarly, our SiT models reuse the same overall Transformer block structure as DiT while incorporating the shift-based operations introduced in the original SiT architecture, and are likewise applied to 1D latent token sequences with absolute positional encodings. Apart from adapting the input interface to our 1D latent representation, we do not introduce any additional architectural modifications to either DiT or SiT.

\paragraph{Evaluation of Representation.}
To measure the representation with semantics and contextual information, we follow previous self-supervised learning~\cite{cvpr2022MAE} and generative models~\cite{cvpr2023MAGE, cvpr2025MergeVQ} to conduct the linear probing protocol, as shown in Table~\ref{tab:in1k_gen_256_vq}. The linear classification is performed upon the latent embedding space of trained encoders by fine-tuning a parameter-free BN layer and a linear layer for 90 epochs using the AdamW optimizer with a batch size of 1024. The basic learning rate is set to $1\times 10^{-3}$ and advanced augmentations and training strategies for modern architectures~\cite{icml2021deit} will not be used.

\paragraph{Evaluation of Generation.}
We evaluate the generative performance of MergeTok-based models on ImageNet-1K in 256$\times$256 resolutions.
As for reconstruction tasks, we follow VQGAN~\cite{cvpr2021vqgan} to report the reconstruction Fr\'{e}chet Inception Distance (rFID) computed on the 50K validation set with \texttt{CenterCrop}. It measures how well the hybrid latent representation preserves the image distribution.
As for class-condition image generation, we strictly follow the setup and use the same reference batches of ADM~\cite{nips2021adm} with their official implementation and evaluate on a single H20-96G GPU and tf32 precision. We report generation FID (gFID), Inception Score (IS), as well as Precision and Recall to jointly characterize fidelity and diversity of the samples. The gFID measures the feature distance between the distributions of real and generated images on the Inception-v3 embeddings under the multivariate Gaussian distributions. IS also uses the Inception-v3 network, but uses the softmax-normalized logit for evaluation of the metric. Unless otherwise specified, all metrics are computed on $50$K generated images and are compared against the ImageNet validation split.
Following common practice~\cite{tian2024VAR}, we report results both with and without classifier-free guidance (CFG). For settings using CFG, we sweep the guidance scale on a held-out subset of the validation set and select the scale that yields the lowest gFID for each generator configuration.

\clearpage
\begin{center}
\rule{0.85\linewidth}{0.4pt}\\[0.4em]
{\large\bfseries Part~IV\;\;|\;\;Extended Experimental Results}\\[0.25em]
{\itshape\small Ablations, codebook diagnostics, and extended evaluations validating MergeTok's design choices.}\\[0.3em]
\rule{0.85\linewidth}{0.4pt}
\end{center}
\vspace{0.4em}

This part presents additional experimental evidence supporting the design choices and performance claims reported in the main paper.
App.~\ref{app:viz} provides further visualizations of reconstruction and generation quality.
App.~\ref{app:ablation} reports additional ablation studies that isolate the contribution of individual components and examine alternative design choices.
App.~\ref{app:exp} extends the comparisons in §\ref{sec:exp_main} with reconstruction and generation experiments at higher resolutions and a cross-domain evaluation on MS-COCO.

\smallskip

\section{Visualization of Analysis}
\label{app:viz}

\begin{figure}[h]
    \centering
    \includegraphics[width=1.0\linewidth]{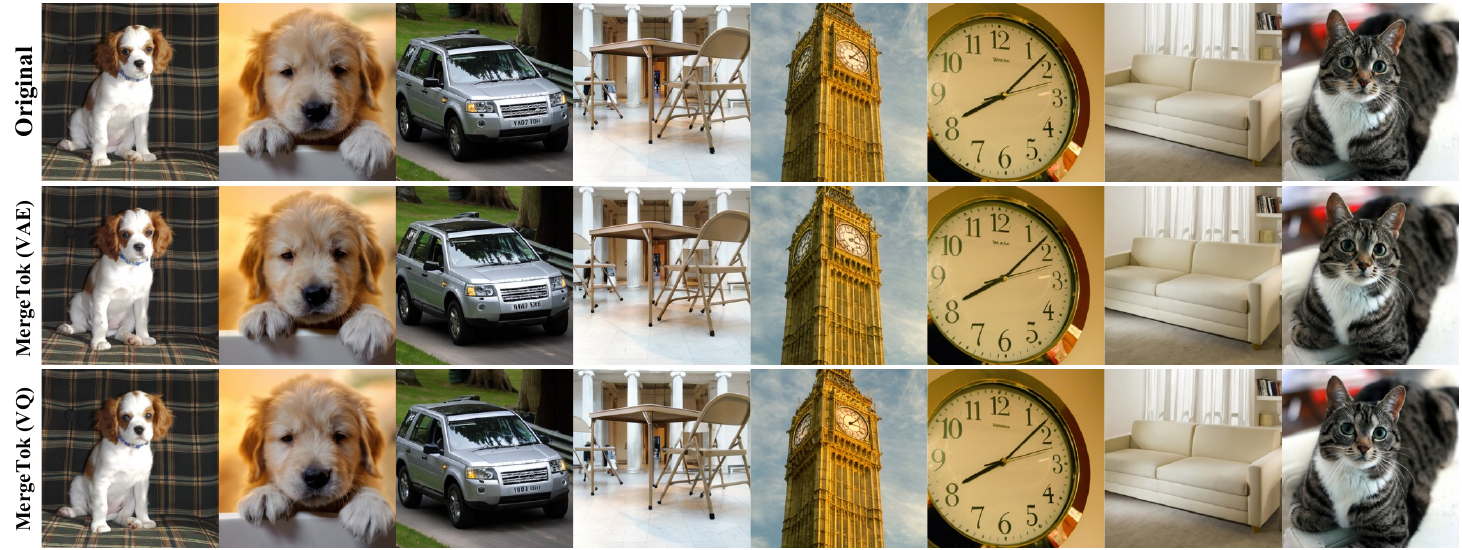}
    \caption{\textbf{Visualization of reconstruction} with the MergeTok-BL tokenizer on ImageNet-256. During inference time, we remove the ToMe modules employed during training from both the VAE and VQ branches to achieve the optimal reconstruction results.
    }
    \label{fig:app_vis_rec_exam}
\end{figure}

\begin{figure}[h]
    \centering
    \includegraphics[width=1.0\linewidth]{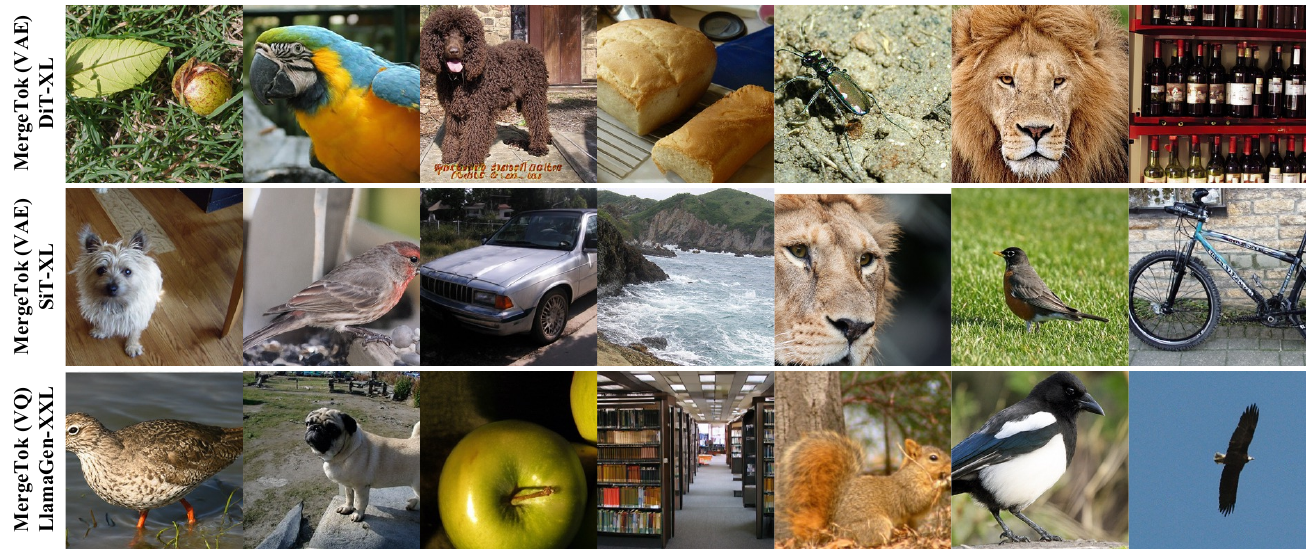}
    \caption{\textbf{Visualization of class-conditional generation} with different generators upon the MergeTok-BL tokenizer in ImageNet-256. Following REPA~\cite{Yu2024REPA} and GigaTok~\cite{iccv2025GigaTok}, we train DiT-XL and SiT-XL generators upon the VAE branch with full tokens (\#256) while training a LlamaGen-XXL generator upon the VQ branch, where these generators achieve competitive generation performances.
    }
    \label{fig:app_vis_gen_exam}
\end{figure}

This section provides qualitative evidence complementing the quantitative comparisons in \S\ref{sec:exp_main}.
Whereas Tables~\ref{tab:in1k_gen_256_vq} and~\ref{tab:in1k_gen_256_vae} report reconstruction and generation metrics, the visualizations here show how the two branches behave at the image level: Fig.~\ref{fig:app_vis_rec_exam} compares VAE and VQ reconstructions from MergeTok-BL on ImageNet-$256$, and Fig.~\ref{fig:app_vis_gen_exam} shows class-conditional generation samples from DiT-XL and SiT-XL on the VAE branch and LlamaGen-XXL on the VQ branch.
We additionally include a brief qualitative analysis of the dual-branch design and its effect on the learned representations.

\paragraph{Visualization of Reconstruction and Generation.} Fig.~\ref{fig:app_vis_rec_exam} shows reconstruction results of both branches of the MergeTok-BL tokenizer on ImageNet-1K at 256$\times$256. Fig.~\ref{fig:app_vis_gen_exam} shows class-conditional generation results using DiT-XL and SiT-XL on the VAE branch and LlamaGen-XXL on the VQ branch.

\paragraph{Experimental Analysis.}
The dual-branch design addresses distinct limitations of VQ and VAE tokenizers. For the VQ branch, continuous gradient updates through the shared encoder alleviate gradient discontinuity in VQ optimization, and the group-constraints loss stabilizes codebook learning. For the VAE branch, semantic alignment with the DINOv2 teacher mitigates the disentanglement issues of standard VAEs while also facilitating more coherent clustering dynamics in the VQ codebook. The qualitative visualizations in Fig.~\ref{fig:rec_vis} support these observations. With semantic alignment, the VAE reconstructions exhibit recognizable structures with clear features. In the VQ branch, the quantized feature maps display stronger semantic interpretability than the original features. Although the codebook lookup introduces slightly denser semantic patterns compared to the VAE branch, the resulting representations retain enhanced recognizability.

\section{More Ablation Studies}
\label{app:ablation}

This section expands the compact ablations in \S\ref{sec:ablation}. Table~\ref{tab:ablation_branch} in the main paper verifies the core design choices using VAE/VQ rFID, whereas the experiments below provide deeper evidence along additional dimensions.
Appendix~\ref{app:ablation_component} performs surgical branch-level ablations on the VQ branch, the VAE branch, and the full joint training in a single unified table.
Appendix~\ref{app:codebook_health} reports codebook-health diagnostics -- usage, \#active, collapse, \#dead, perplexity, and entropy -- to characterize VQ stability beyond rFID.
Appendix~\ref{app:multi_metric} evaluates reconstruction, representation, and generation jointly (rFID, PSNR, SSIM, linear probing accuracy, AR gFID, and diffusion gFID).
Appendix~\ref{app:alt_grouping} tests whether the ToMe-derived source map is preferable to alternative grouping strategies of matched group count.

\subsection{Surgical Component Ablation}
\label{app:ablation_component}

We isolate the contribution of each component in MergeTok by ablating the VQ branch, the VAE branch, and the full joint training separately. All experiments use the SB configuration.

\begin{table}[h]
\centering
\caption{\textbf{Surgical component ablations on MergeTok-SB.}
We ablate the VQ branch and the VAE branch separately,
then enable both with the full joint training.
\textbf{ToMe}: token merging in the encoder;
\textbf{Align.}: matched-granularity alignment with DINOv2;
\textbf{Cons.}: group consistency loss on the VQ codebook.}
\label{tab:ablation_components}
\setlength{\tabcolsep}{6pt}
\small
\begin{tabular}{l|ccc|cc}
\toprule
Variant & ToMe & Align. & Cons. & VQ-rFID$\downarrow$ & VAE-rFID$\downarrow$ \\
\midrule
\multicolumn{6}{l}{\cellcolor{gray!10}\textit{(a) VQ branch only}} \\
\midrule
VQ Baseline (GigaTok-SB w/o Dist) &  --        &  --        &  --        & 1.12 & --   \\
\brow VQ + Consistency Loss       &  --        &  --        & \checkmark & 1.04 & --   \\
\midrule
\multicolumn{6}{l}{\cellcolor{gray!10}\textit{(b) VAE branch only}} \\
\midrule
VAE Baseline                       &  --        &  --        &  --        & --   & 0.81 \\
VAE + ToMe                         & \checkmark &  --        &  --        & --   & 0.67 \\
VAE + Alignment                    &  --        & \checkmark &  --        & --   & 0.63 \\
\brow VAE + ToMe + Alignment       & \checkmark & \checkmark &  --        & --   & 0.59 \\
\midrule
\multicolumn{6}{l}{\cellcolor{gray!10}\textit{(c) Full joint training}} \\
\midrule
\brow MergeTok-SB                  & \checkmark & \checkmark & \checkmark & 0.96 & 0.59 \\
\bottomrule
\end{tabular}
\end{table}

Block (a) of Table~\ref{tab:ablation_components} shows that adding the group consistency loss alone improves VQ-rFID from 1.12 to 1.04. In this variant, the source map $S$ is extracted via a double-forward pass without the VAE branch. The result confirms that group-aware regularization is beneficial on its own, although the gains remain moderate without the continuous gradient flow provided by joint training with the VAE branch. Block (b) isolates the VAE branch: token merging is the primary contributor, reducing VAE-rFID from 0.81 to 0.67, and matched-granularity alignment provides a further reduction to 0.59 by aligning the merged tokens with DINOv2 semantic features. Block (c) enables both branches with full joint training and achieves the best result on both branches simultaneously. The VQ branch improves from 1.12 to 0.96 even though the VQ forward path itself does not use ToMe or alignment. We attribute this gain to the continuous gradient flow from the VAE branch through the shared encoder. The VAE branch achieves 0.59 in both standalone and dual-branch settings, indicating that it operates near its architectural ceiling and primarily serves as a structural prior provider for the VQ branch during joint training. This branch-level decomposition supports the component-integration trend reported in Table~\ref{tab:ablation_branch_a} of the main paper.

\subsection{Codebook Health Diagnostics}
\label{app:codebook_health}

We complement the reconstruction FID evaluation with direct codebook health metrics to validate the VQ stabilization effect. All metrics are measured at convergence on the ImageNet-1K validation set with codebook size 16{,}384.

\begin{table}[h]
    \centering
    \caption{\textbf{Codebook health under different training regimes.}
    Codebook size is 16{,}384.
    \#Active~$=$~Usage~$\times$~16{,}384;
    Entropy~$=\log_2(\text{Perplexity})$;
    rFID is the corresponding VQ reconstruction quality.}
    \label{tab:codebook_health}
    \setlength{\tabcolsep}{3.5pt}
    \small
    \begin{tabular}{l|cc|cc|cc|c}
    \toprule
    Method
      & Usage(\%)$\uparrow$ & \#Active$\uparrow$
      & Collapse(\%)$\downarrow$ & \#Dead$\downarrow$
      & Perplexity$\uparrow$ & Entropy(bits)$\uparrow$
      & rFID$\downarrow$ \\ \midrule
    VQ Baseline          & 72 & 11{,}796 & 28.0 & 4{,}588 & 3{,}840 & 11.91 & 1.12 \\
    VQ + Cons.\ Loss     & 81 & 13{,}271 & 19.0 & 3{,}113 & 5{,}120 & 12.32 & 1.04 \\
    \brow MergeTok-SB    & 93 & 15{,}237 &  7.0 & 1{,}147 & 9{,}800 & 13.26 & 0.96 \\
    \bottomrule
    \end{tabular}
\end{table}

Codebook utilization improves by 21 percentage points from 72\% to 93\%, the collapse rate drops from 28\% to 7\%, and perplexity increases by approximately 2.5$\times$. These measurements provide direct empirical support for the mechanism described in Sec.~\ref{sec3.4_vq}, confirming that group-aware regularization combined with continuous gradient flow jointly improves codebook learning. This provides mechanism-level evidence for the VQ improvements discussed in \S\ref{sec:ablation}.

\subsection{Multi-Metric and Multi-Task Evaluation}
\label{app:multi_metric}

To verify that the benefit of ToMe generalizes beyond rFID, we evaluate all configurations across reconstruction quality measured by rFID, PSNR, and SSIM, representation quality measured by linear probing accuracy, and downstream generation measured by AR and diffusion gFID.

\begin{table}[h]
    \centering
    \caption{\textbf{Multi-metric ablation on MergeTok-SB.} AR gFID is measured with LlamaGen-B at 111M parameters trained for 300 epochs. Diffusion gFID is measured with SiT-L at 458M parameters trained for 400K steps with CFG. Full-scale VAE generation with SiT-XL achieves gFID 1.79 as reported in Table~2.}
    \setlength{\tabcolsep}{3pt}
    \small
    \begin{tabular}{l|ccc|c|cc}
    \toprule
    Setting & rFID$\downarrow$ & PSNR$\uparrow$ & SSIM$\uparrow$ & Lin.Acc$\uparrow$ & AR gFID$\downarrow$ & Diff.~gFID$\downarrow$ \\ \midrule
    VQ Baseline & 1.12 & 20.5 & 0.665 & 61.5 & 3.83 & --- \\
    VAE Baseline & 0.81 & 21.4 & 0.680 & 55.2 & --- & 8.14 \\
    VAE + ToMe & 0.75 & 21.6 & 0.686 & 58.4 & --- & 6.87 \\
    VAE + ToMe + Align. & 0.59 & 21.9 & 0.693 & 66.1 & --- & 5.43 \\
    \brow MergeTok-SB (VQ/VAE) & 0.96\,/\,0.59 & 21.1\,/\,21.9 & 0.678\,/\,0.693 & 73.8 & 3.37 & 4.96 \\
    \bottomrule
    \end{tabular}
    \label{tab:multi_metric}
\end{table}

The improvements are consistent across all metrics. ToMe alone improves PSNR by 0.2\,dB and linear probing accuracy by 3.2 points. Full MergeTok-SB achieves a 12.3-point linear probing accuracy gain over the VQ baseline and reduces AR gFID from 3.83 to 3.37 with the same generator. This confirms that the gains in \S\ref{sec:ablation} extend beyond rFID to representation and generation.

\subsection{Alternative Grouping Strategies}
\label{app:alt_grouping}

To isolate the contribution of ToMe's hierarchical, attention-aware clustering from the more general benefit of having any group prior, we replace the ToMe-derived source map $S$ with four alternative grouping strategies of matched group count $K=128$ and retrain MergeTok-SB under identical hyperparameters. The five strategies are:
\begin{enumerate}
    \item \textbf{ToMe}: the default attention-based, layer-wise merging.
    \item \textbf{K-means on encoder output}: offline k-means clustering applied to the final encoder output $Z_L$, run once at each training step on the current batch.
    \item \textbf{K-means on frozen DINOv2 features}: clustering on the teacher's patch features rather than the student's.
    \item \textbf{Spatial grid grouping}: a fixed $16{\times}16 \to 8{\times}8$ spatial pooling that ignores content and groups tokens purely by image position.
    \item \textbf{Random partition}: tokens are randomly assigned to $K$ groups with a fixed seed per image.
\end{enumerate}
All five variants share the same VAE and VQ branches, alignment loss, and group-aware losses $\mathcal{L}_{\mathrm{div}}$ and $\mathcal{L}_{\mathrm{cons}}$. Only the source of $S$ differs.

\begin{table}[h]
    \centering
    \caption{\textbf{Impact of alternative grouping strategies on MergeTok-SB.} All variants share the same VAE/VQ branches, alignment loss, and group-aware losses; only the source of the source map $S$ differs.}
    \setlength{\tabcolsep}{3pt}
    \small
    \begin{tabular}{l|cc|c|cc}
    \toprule
    Strategy & VAE-rFID$\downarrow$ & VQ-rFID$\downarrow$ & Lin.Acc$\uparrow$ & Usage(\%)$\uparrow$ & Collapse(\%)$\downarrow$ \\ \midrule
    No grouping & 0.81 & 1.12 & 61.5 & 72 & 28 \\
    Random partition & 0.84 & 1.10 & 57.9 & 78 & 22 \\
    Spatial grid & 0.72 & 1.03 & 64.7 & 85 & 16 \\
    K-means on encoder & 0.66 & 1.00 & 70.1 & 88 & 12 \\
    K-means on DINOv2 & 0.63 & 0.99 & 71.6 & 90 & 10 \\
    \brow ToMe (Ours) & \textbf{0.59} & \textbf{0.96} & \textbf{73.8} & \textbf{93} & \textbf{7} \\
    \bottomrule
    \end{tabular}
    \label{tab:alt_grouping}
\end{table}

We summarize the findings from Table~\ref{tab:alt_grouping} below.

\textit{Content-aware grouping consistently helps.} All content-aware strategies, including ToMe and both k-means variants, outperform the no-grouping baseline on every metric, confirming that structural priors stabilize both VQ codebook learning and VAE alignment. The gap between ToMe and k-means on encoder features is nonetheless substantial, with VAE-rFID of 0.59 versus 0.66 and linear probing accuracy of 73.8 versus 70.1. This indicates that the layer-wise adaptive clustering that co-evolves with training is more effective than a one-shot post-hoc clustering.

\textit{Teacher-based clustering outperforms student-based clustering.} K-means on DINOv2 features achieves 0.63 VAE-rFID compared with 0.66 for k-means on encoder features, suggesting that the frozen teacher provides a more stable clustering target. ToMe still surpasses both variants because it operates inside each attention block and co-adapts with the encoder across layers.

\textit{Random partitions degrade performance.} Random grouping yields 0.84 VAE-rFID, 1.10 VQ-rFID, and 57.9 linear probing accuracy, which are worse than the no-grouping baseline on the latter two metrics. When $S$ is noise, $\mathcal{L}_{\mathrm{cons}}$ enforces spurious code exclusivity across unrelated groups and $\mathcal{L}_{\mathrm{div}}$ encourages diversity within meaningless clusters, disrupting codebook learning rather than regularizing it.

\textit{Spatial grid grouping falls between content-aware and random strategies.} It achieves 0.72 VAE-rFID and 64.7 linear probing accuracy. The fixed spatial pooling captures coarse locality but ignores semantic content, which limits both alignment quality and codebook organization.

Together, these comparisons support the use of the ToMe-derived source map in the main design and are consistent with the alternative-grouping claim made in \S\ref{sec:ablation}.

\section{More Results on Benchmarks}
\label{app:exp}

This section extends the main comparisons in \S\ref{sec:exp_main} beyond the default ImageNet-$256$ benchmark. We evaluate whether the reconstruction and generation advantages observed in Tables~\ref{tab:in1k_gen_256_vq} and~\ref{tab:in1k_gen_256_vae} persist at higher resolutions and under a cross-domain setting on MS-COCO. These experiments are intended as generalization checks rather than new method variants: the tokenizer and generator architectures are unchanged.
Appendix~\ref{app:higher_res_rec} evaluates reconstruction quality at $512{\times}512$ and $1024{\times}1024$ on ImageNet, examining zero-shot generalization to unseen resolutions.
Appendix~\ref{app:higher_res_gen} evaluates class-conditional generation at $512{\times}512$ on ImageNet.
Appendix~\ref{app:coco} provides a cross-domain reconstruction evaluation on the MS-COCO 2017 validation set to assess whether the improvements observed on ImageNet hold under more heterogeneous scene distributions.

\subsection{Higher Resolution Reconstruction}
\label{app:higher_res_rec}

We evaluate MergeTok's reconstruction quality at $512{\times}512$ and $1024{\times}1024$ on ImageNet to examine zero-shot generalization beyond the training resolution. The $256{\times}256$ checkpoint is applied directly by tiling the positional embeddings, with no architectural modification or fine-tuning.

\begin{table}[ht]
\centering
\caption{\textbf{Reconstruction on ImageNet at higher resolutions.}
Top block: 512$\times$512. Bottom block: 1024$\times$1024.
All MergeTok results use the 256$\times$256 checkpoint via
positional-embedding tiling, with no fine-tuning.}
\label{tab:rec_imagenet_highres}
\setlength{\tabcolsep}{4pt}
\small
\begin{tabular}{l|cccc|ccc}
\toprule
Tokenizer & Type & Ratio & \#Tok. & \#Code & rFID$\downarrow$ & PSNR$\uparrow$ & SSIM$\uparrow$ \\
\midrule
\multicolumn{8}{l}{\cellcolor{gray!10}\textit{Resolution: 512$\times$512}} \\
\midrule
VAE~{\tiny\cite{iclr2013VAE}}                         & 2D VAE    & 8  & $32^2$ & --       & 0.62          & \bf{24.3}     & \bf{0.71}     \\
SoftVQ-BL~{\tiny\cite{cvpr2025SoftVQVAE}}             & 1D VQ     & 32 & 64     & --       & 0.71          & --            & --            \\
\brow \bf{MergeTok-BL (VAE)}                          & 1D VAE+VQ & 16 & 256    & $2^{14}$ & \bf{0.42}     & 21.8          & 0.69          \\
\brow \bf{MergeTok-BL (VQ)}                           & 1D VAE+VQ & 16 & 256    & $2^{14}$ & 0.82          & 21.0          & 0.67          \\
\midrule
\multicolumn{8}{l}{\cellcolor{gray!10}\textit{Resolution: 1024$\times$1024}} \\
\midrule
\small{Taming-VQGAN}~{\tiny\cite{cvpr2021vqgan}}      & 2D VQ     & 16 & $16^2$ & $2^{10}$ & 6.02          & 19.20         & 0.60          \\
OpenMAGVIT2~{\tiny\cite{luo2024Open-Magvit2}}         & 2D LFQ    & 16 & $16^2$ & $2^{18}$ & 3.43          & 19.45         & 0.63          \\
LlamaGen-Tok.~{\tiny\cite{NIPS2024LLaMAGen}}          & 2D VQ     & 32 & $8^2$  & $2^{14}$ & 3.17          & 19.94         & 0.64          \\
TiTok-S-128~{\tiny\cite{nips2024titok}}               & 1D VQ     & 32 & 128    & $2^{12}$ & 2.32          & 16.97         & 0.51          \\
Layton-H                                              & 1D Diff.  & 32 & --     & --       & 2.78          & 19.80         & 0.65          \\
\brow \bf{MergeTok-BL (VQ)}                           & 1D VAE+VQ & 16 & 256    & $2^{14}$ & \bf{1.87}     & \bf{20.14}    & \bf{0.67}     \\
\bottomrule
\end{tabular}
\end{table}

The top block of Table~\ref{tab:rec_imagenet_highres} reports the $512{\times}512$ setting. The VAE branch of MergeTok-BL achieves rFID 0.42, outperforming the SD-VAE baseline (0.62) and SoftVQ-BL (0.71). The VQ branch reaches rFID 0.82; this is competitive given that 1D tokenizers face a stronger positional extrapolation challenge at unseen resolutions than their 2D counterparts.
The bottom block of Table~\ref{tab:rec_imagenet_highres} extends the evaluation to $1024{\times}1024$, where the margins widen. MergeTok-BL (VQ) achieves rFID 1.87, a 19\% relative reduction over TiTok-S-128 (2.32), the strongest discrete baseline, and leads on both PSNR (20.14 vs.\ 19.94) and SSIM (0.67 vs.\ 0.64 for LlamaGen-Tok.). The consistent resolution-scaling behavior suggests that the attention-key similarity criterion underlying ToMe produces groupings that remain semantically coherent as the token sequence scales with image size.

\begin{table}[ht]
    \centering
    \caption{\textbf{Generation on ImageNet 512$\times$512.}}
    \setlength{\tabcolsep}{4pt}
    \small
    \begin{tabular}{l|ccc|cc|cc}
    \toprule
    Tokenizer & Type & Ratio & \#Tok. & Generator & \#Param. & gFID$\downarrow$ & IS$\uparrow$ \\
    \midrule
    SoftVQ-BL~{\tiny\cite{cvpr2025SoftVQVAE}}  & 1D VQ     & 16 & 64    & SiT-XL      & 675M & 7.96 & 133.9 \\
    MaskGIT-VQ~{\tiny\cite{CVPR2022maskgit}}   & 2D VQ     & 16 & $16^2$& MaskGIT-ViT & 177M & 3.72 & 156.0 \\
    TiTok-B-64~{\tiny\cite{nips2024titok}}     & 1D VQ     & 16 & 64    & MaskGIT-ViT & 177M & 3.64 & 179.8 \\
    TiTok-L-32~{\tiny\cite{nips2024titok}}     & 1D VQ     & 16 & 32    & MaskGIT-ViT & 177M & 3.91 & 182.0 \\
    SoftVQ-BL~{\tiny\cite{cvpr2025SoftVQVAE}}  & 1D VQ     & 16 & 64    & MAR-H       & 479M & 8.21 & 152.9 \\
    \brow \bf{MergeTok (VQ)} & 1D VAE+VQ & 16 & 256 & LlamaGen & 343M & \bf{3.23} & \bf{217.4} \\
    \bottomrule
    \end{tabular}
    \label{tab:gen_imagenet_512}
\end{table}

\begin{table}[ht]
    \centering
    \caption{\textbf{Reconstruction on MS-COCO 2017 val.}}
    \setlength{\tabcolsep}{4pt}
    \small
    \begin{tabular}{l|ccc|cccc}
    \toprule
    Method & Type & \#Tok. & \#Code & Res. & rFID$\downarrow$ & PSNR$\uparrow$ & SSIM$\uparrow$ \\
    \midrule
    GigaTok-SB~{\tiny\cite{iccv2025GigaTok}}        & 1D VQ     & 256 & $2^{14}$ & 256 & 3.8          & 20.9          & 0.66          \\
    GigaTok-BL~{\tiny\cite{iccv2025GigaTok}}        & 1D VQ     & 256 & $2^{14}$ & 256 & 2.6          & 21.5          & 0.68          \\
    \brow \bf{MergeTok-BL (VAE)}                    & 1D VAE+VQ & 256 & $2^{14}$ & 256 & \bf{1.8}     & \bf{22.3}     & \bf{0.70}     \\
    \brow \bf{MergeTok-BL (VQ)}                     & 1D VAE+VQ & 256 & $2^{14}$ & 256 & 2.4          & 21.7          & 0.68          \\
    \midrule
    GigaTok-BL~{\tiny\cite{iccv2025GigaTok}}        & 1D VQ     & 256 & $2^{14}$ & 512 & 4.2          & 21.9          & 0.69          \\
    \brow \bf{MergeTok-BL (VAE)}                    & 1D VAE+VQ & 256 & $2^{14}$ & 512 & \bf{2.9}     & \bf{22.8}     & \bf{0.71}     \\
    \brow \bf{MergeTok-BL (VQ)}                     & 1D VAE+VQ & 256 & $2^{14}$ & 512 & 3.6          & 22.1          & 0.69          \\
    \bottomrule
    \end{tabular}
    \label{tab:coco_rec}
\end{table}

\subsection{Higher Resolution Generation}
\label{app:higher_res_gen}

We evaluate class-conditional image synthesis at $512{\times}512$ on ImageNet to assess whether the reconstruction gains described above carry over to generation quality.

Table~\ref{tab:gen_imagenet_512} presents the results. MergeTok with LlamaGen achieves gFID 3.23 and IS 217.4. Among masked-prediction generators, this represents improvements of 0.41 in gFID over TiTok-B-64 and 0.68 over TiTok-L-32, with an IS gain of 35.4 points over the latter. Among flow-based generators, SoftVQ-BL with SiT-XL records gFID 7.96, a gap of 4.73 relative to our result. The gains on both gFID and IS suggest that the improved codebook utilization reported in App.~\ref{app:codebook_health} translates into more faithful and diverse generated samples at higher resolution.

\subsection{MS-COCO Cross-Domain Evaluation}
\label{app:coco}

The ImageNet benchmark comprises predominantly object-centric, single-label images with relatively uncluttered backgrounds.
To assess whether the observed improvements extend to more challenging scene distributions, we evaluate reconstruction quality on the MS-COCO 2017 validation set~\cite{Lin2014COCO}, which contains multi-object scenes with varied spatial configurations, overlapping instances, and richer background context.
No domain adaptation or fine-tuning is performed; the same checkpoints trained on ImageNet are evaluated directly on COCO.

Table~\ref{tab:coco_rec} reports results at $256{\times}256$ and $512{\times}512$.
At $256{\times}256$, the VAE branch of MergeTok-BL achieves rFID 1.8, a 31\% relative reduction over GigaTok-BL (2.6); correspondingly, PSNR improves from 21.5 to 22.3\,dB and SSIM from 0.68 to 0.70.
The VQ branch likewise surpasses GigaTok-BL, with rFID 2.4 versus 2.6.
At $512{\times}512$, the same ordering is preserved: the VAE branch attains rFID 2.9 against 4.2 for GigaTok-BL, a 31\% relative improvement consistent with the $256{\times}256$ result.
The uniform gains across both resolutions and both branches indicate that the content-adaptive token grouping induced by ToMe captures structural regularity that is not limited to the single-label distribution of ImageNet, but generalizes to the more heterogeneous scene structure present in COCO.

\end{document}